\documentclass[11pt,a4paper]{article}

\usepackage{amsmath,amssymb,amsthm}

\usepackage{graphicx}
\usepackage{booktabs}
\usepackage{multirow}
\usepackage{subcaption}

\usepackage{xcolor}

\usepackage[ruled,vlined,linesnumbered]{algorithm2e}

\usepackage{listings}
\usepackage{hyperref}
\hypersetup{
  colorlinks=true,
  linkcolor=blue,
  citecolor=blue,
  urlcolor=blue
}

\usepackage[margin=2.5cm]{geometry}

\usepackage{enumitem}
\usepackage{float}

\newtheorem{definition}{Definition}

\title{\textbf{cuGenOpt: A GPU-Accelerated General-Purpose Metaheuristic Framework\\for Combinatorial Optimization}}

\author{
  Yuyang Liu \\
  Independent Researcher, Shenzhen, China \\
  \texttt{15251858055@163.com}
}

\date{}

\begin{document}

\maketitle

\begin{abstract}
Combinatorial optimization problems arise in logistics, scheduling, and resource allocation,
yet existing approaches face a fundamental trade-off among generality, performance, and usability.
Exact methods scale poorly beyond moderate sizes;
specialized solvers achieve strong results but only on narrow problem classes;
CPU-based metaheuristics offer generality but limited throughput.
We present \textbf{cuGenOpt}, a GPU-accelerated general-purpose metaheuristic framework
that addresses all three dimensions simultaneously.

At the \emph{engine} level, cuGenOpt adopts a ``one block evolves one solution'' CUDA architecture
with a unified encoding abstraction (permutation, binary, integer),
a two-level adaptive operator selection (AOS) mechanism,
and hardware-aware resource management
(shared-memory auto-extension, L2-cache-aware population sizing).
At the \emph{extensibility} level, a user-defined operator registration interface
allows domain experts to inject problem-specific CUDA search operators
that participate in AOS weight competition alongside built-in operators,
bridging the gap between generality and specialization.
At the \emph{usability} level, a JIT compilation pipeline
exposes the entire framework as a pure-Python API (\texttt{pip install cugenopt}),
and an LLM-based modeling assistant converts natural-language problem descriptions
into executable solver code without manual CUDA programming.

Experiments across five thematic suites on three GPU architectures (T4, V100, A800) show that
cuGenOpt outperforms general MIP solvers by orders of magnitude,
achieves competitive quality against specialized solvers on instances up to $n{=}150$,
and attains 4.73\% gap on TSP-442 within 30\,s on A800.
Twelve problem types spanning five encoding variants are solved to optimality,
and standard benchmarks (QAPLIB, Solomon VRPTW, OR-Library JSP) confirm near-optimal convergence.
Framework-level optimizations cumulatively reduce pcb442 gap from 36\% to 4.73\%
and boost VRPTW throughput by 75--81\% via shared-memory extension alone.

\noindent\textbf{Code:} \url{https://github.com/L-yang-yang/cugenopt}
\end{abstract}

\vspace{0.5em}

\section{Introduction}
\label{sec:intro}

Combinatorial optimization problems pervade logistics, scheduling, resource allocation,
and numerous other domains.
Classical problems such as the Traveling Salesman Problem (TSP),
Vehicle Routing Problem (VRP), Job-Shop Scheduling (JSP),
and Quadratic Assignment Problem (QAP) are NP-hard,
and the computational cost of exact solutions grows exponentially
with problem size~\cite{papadimitriou1998combinatorial}.

Current approaches fall into three broad categories,
each facing distinct limitations:

\begin{enumerate}[leftmargin=*]
  \item \textbf{Exact methods} (e.g., Mixed-Integer Programming):
    Guarantee global optimality through mathematical modeling.
    However, general-purpose MIP formulations such as the MTZ model
    for TSP~\cite{miller1960integer} introduce $O(n^2)$ variables;
    solvers like SCIP~\cite{scip} often fail to find high-quality
    feasible solutions within reasonable time when $n > 100$.
    Moreover, MIP modeling demands operations research expertise,
    particularly for nonlinear objectives and complex constraints.

  \item \textbf{Specialized solvers} (e.g., OR-Tools Routing~\cite{ortools},
    NVIDIA cuOpt~\cite{cuopt}):
    Highly optimized for specific problem classes with excellent performance,
    but limited in scope---they cannot accommodate custom constraints
    (e.g., intra-route priority ordering) or nonlinear objective functions,
    and require ground-up development for new problem types.

  \item \textbf{Metaheuristics} (e.g., simulated annealing~\cite{kirkpatrick1983optimization},
    evolutionary algorithms~\cite{eiben2015introduction}):
    Highly general, applicable to arbitrary black-box objectives.
    Traditional CPU implementations, however, offer limited search throughput
    and converge slowly on large-scale instances.
    GPU-accelerated metaheuristics~\cite{luong2013gpu,cecilia2013enhancing}
    achieve significant speedups but are typically designed for a single problem type,
    lacking a general-purpose framework abstraction.
\end{enumerate}

An underserved space lies at the intersection of these three categories:
\textbf{combining GPU-accelerated search throughput,
general-purpose problem modeling,
and a low-barrier user interface}.
This paper presents cuGenOpt to fill that gap.

\subsection{Overview of cuGenOpt}

cuGenOpt is a GPU-accelerated general-purpose metaheuristic framework
for combinatorial optimization, organized around three progressive layers:

\textbf{Core engine} (Sections~\ref{sec:framework}--\ref{sec:adaptive}).
The framework adopts a ``one block evolves one solution'' CUDA parallel architecture
supporting three universal encoding types: permutation, binary, and integer.
Search strategy is dynamically adjusted through a two-level adaptive operator selection (AOS) mechanism,
with problem-profile-driven prior weights accelerating convergence.
Hardware-aware mechanisms---shared-memory auto-extension
and L2-cache-aware adaptive population sizing---enable the same binary
to automatically select optimal memory paths across different GPUs.

\textbf{Extensibility} (Section~\ref{sec:framework}).
A user-defined operator registration interface allows domain experts
to inject problem-specific CUDA search operators
(e.g., delta-evaluation 2-opt for TSP) as code snippets.
These operators are JIT-compiled into the framework's operator system
and compete with built-in operators through AOS weight adaptation,
providing a specialization channel while preserving framework generality.

\textbf{User interface} (Section~\ref{sec:interface}).
A JIT compilation pipeline packages the entire framework
as a pure-Python package (\texttt{pip install cugenopt}),
enabling users to solve built-in problems or define custom problems
through code snippets without writing full CUDA programs.
An LLM-based modeling assistant further converts
natural-language problem descriptions into executable solver code.

\subsection{Contributions}

\begin{enumerate}[leftmargin=*]
  \item \textbf{General-purpose GPU solver framework}:
    We propose a ``one block evolves one solution'' GPU parallel architecture
    with unified support for permutation, binary, and integer encodings,
    validated on 12 representative problem types.
    User modeling effort is typically 20--50 lines of CUDA.

  \item \textbf{Multi-layer adaptive search with hardware awareness}:
    We design problem-profile-driven operator prior weights (L1)
    and two-level EMA-based adaptive operator selection (L3),
    coupled with shared-memory auto-extension
    and L2-cache-aware population sizing
    for joint optimization of search strategy and GPU resources.

  \item \textbf{Extensible operator system}:
    We propose a user-defined operator registration mechanism
    using type-erased data passing and JIT injection,
    enabling user-written CUDA operators to seamlessly integrate
    into the AOS framework.

  \item \textbf{End-to-end usability}:
    A JIT compilation pipeline delivers a pure-Python API,
    and an LLM-based modeling assistant supports
    natural-language-to-GPU-solving workflows,
    reducing the barrier from CUDA programming to Python function calls.

  \item \textbf{Comprehensive experimental validation}:
    Five thematic experiment suites on three GPU architectures
    (T4, V100, A800) evaluate baseline comparisons,
    scalability, generality, optimization ablation,
    and user-defined operator effectiveness.
\end{enumerate}

\subsection{Paper Organization}

Section~\ref{sec:related} reviews related work.
Section~\ref{sec:framework} presents the framework design
including user-defined operator registration.
Section~\ref{sec:adaptive} details adaptive search and hardware-aware mechanisms.
Section~\ref{sec:interface} describes the user interface and toolchain.
Section~\ref{sec:experiments} presents experimental results.
Section~\ref{sec:discussion} discusses key findings and limitations.
Section~\ref{sec:conclusion} concludes the paper.

\section{Related Work}
\label{sec:related}

\subsection{GPU-Accelerated Metaheuristics}

The massively parallel architecture of GPUs is a natural fit
for population-based metaheuristics.
Luong et al.~\cite{luong2013gpu} systematically categorize
three parallelization modes for GPU-based local search:
solution-level, move-level, and iteration-level parallelism.
Cecilia et al.~\cite{cecilia2013enhancing} port ant colony optimization to GPUs,
exploiting data parallelism for pheromone updates and path construction.
Delevacq et al.~\cite{delevacq2013parallel} implement iterated local search
for TSP on GPUs.
Zhou and Tan~\cite{zhou2018gpu} propose a GPU-accelerated evolutionary
computation framework, though its design centers on genetic algorithms
with limited support for local search operators.

A common limitation of these works is \textbf{problem specificity}:
each targets a particular problem (typically TSP or continuous optimization),
requiring users to re-implement parallel evaluation and neighborhood search
for every new problem.

\subsection{General-Purpose Combinatorial Optimization Frameworks}

On the CPU side, Google OR-Tools~\cite{ortools} provides MIP solver interfaces
and a specialized Routing solver based on Guided Local Search (GLS),
achieving strong performance on TSP/VRP but unable to express
custom semantics such as intra-route priority constraints.
Ropke and Pisinger~\cite{ropke2006adaptive} propose Adaptive Large Neighborhood Search (ALNS),
offering flexible neighborhood structures through destroy-repair operator pairs,
but its single-threaded CPU implementation limits throughput on large instances,
and each new problem requires custom operator design.

On the GPU side, NVIDIA cuOpt~\cite{cuopt} is a commercial-grade
GPU-accelerated solver covering MILP/LP and vehicle routing,
achieving state-of-the-art performance on standard benchmarks.
However, its problem scope is fixed:
the routing module supports only predefined VRP constraint types,
the MILP module requires explicit linear formulations,
and users cannot define arbitrary black-box objectives.

\subsection{Adaptive Operator Selection}

Adaptive Operator Selection (AOS) is an active research area in metaheuristics.
Fialho et al.~\cite{fialho2010adaptive} model AOS as a dynamic multi-armed bandit (MAB) problem,
balancing exploration and exploitation via UCB strategies.
Li et al.~\cite{li2014adaptive} apply MAB-based AOS to MOEA/D.
Maturana and Saubion~\cite{maturana2009autonomous} propose autonomous operator management
through credit assignment and probability matching.

cuGenOpt's AOS innovates in three aspects:
(1)~\textbf{two-level adaptation}---adjusting weights at both
the operator (Sequence) level and the search step-count (K-step) level;
(2)~\textbf{GPU-native implementation}---statistics are collected
via shared-memory atomic operations entirely on the GPU;
(3)~\textbf{problem-profile priors}---static analysis of problem structure
sets initial operator weights before AOS begins runtime adaptation.

\subsection{JIT Compilation in Scientific Computing}

Just-In-Time (JIT) compilation is widely used in scientific computing.
Numba~\cite{lam2015numba} compiles Python numerical code to machine code via LLVM.
CuPy~\cite{cupy} provides a NumPy-compatible GPU array interface
with support for user-defined CUDA kernels.
PyCUDA~\cite{pycuda} enables dynamic generation and compilation of CUDA code from Python.
cuGenOpt's JIT pipeline differs in \textbf{compilation granularity}:
rather than compiling individual kernels or array operations,
it embeds user-provided problem definitions (objectives, constraints, data)
into a complete solver template and compiles a standalone executable,
achieving runtime performance equivalent to hand-written CUDA
while maintaining a clean Python interface.

\subsection{Positioning of This Work}

Table~\ref{tab:comparison} summarizes the comparison between cuGenOpt and existing approaches.
cuGenOpt is, to our knowledge, the first combinatorial optimization framework
that simultaneously provides \textbf{GPU parallelism},
\textbf{universal encodings}, \textbf{adaptive operator selection},
\textbf{user operator extensibility}, and a \textbf{Python API}.

\begin{table}[htbp]
\centering
\caption{Feature comparison of cuGenOpt with existing approaches.}
\label{tab:comparison}
\small
\begin{tabular}{lcccccc}
\toprule
\textbf{Method} & \textbf{GPU} & \textbf{Univ.\ Enc.} & \textbf{AOS} & \textbf{Op.\ Ext.} & \textbf{Python} & \textbf{New Problem} \\
\midrule
MIP (SCIP/CBC)          & $\times$ & $\times$ & $\times$ & $\times$ & $\checkmark$ & Math modeling \\
OR-Tools Routing        & $\times$ & $\times$ & $\times$ & $\times$ & $\checkmark$ & Not supported \\
cuOpt~\cite{cuopt}      & $\checkmark$ & $\times$ & $\times$ & $\times$ & $\checkmark$ & VRP/MILP only \\
GPU-ACO~\cite{cecilia2013enhancing} & $\checkmark$ & $\times$ & $\times$ & $\times$ & $\times$ & Rewrite kernel \\
ALNS~\cite{ropke2006adaptive}       & $\times$ & $\checkmark$ & $\checkmark$ & $\checkmark$ & $\times$ & Custom op.\ pairs \\
GPU-EA~\cite{zhou2018gpu}           & $\checkmark$ & Partial & $\times$ & $\times$ & $\times$ & Partial rewrite \\
\midrule
\textbf{cuGenOpt}       & $\checkmark$ & $\checkmark$ & $\checkmark$ & $\checkmark$ & $\checkmark$ & \textbf{20--50 lines} \\
\bottomrule
\end{tabular}
\end{table}

\section{Framework Design}
\label{sec:framework}

This section presents the core architecture of cuGenOpt,
covering problem abstraction, GPU parallelization strategy,
the search operator system (including user-defined operator registration),
and population management.

\begin{figure}[htbp]
\centering
\includegraphics[width=0.95\columnwidth]{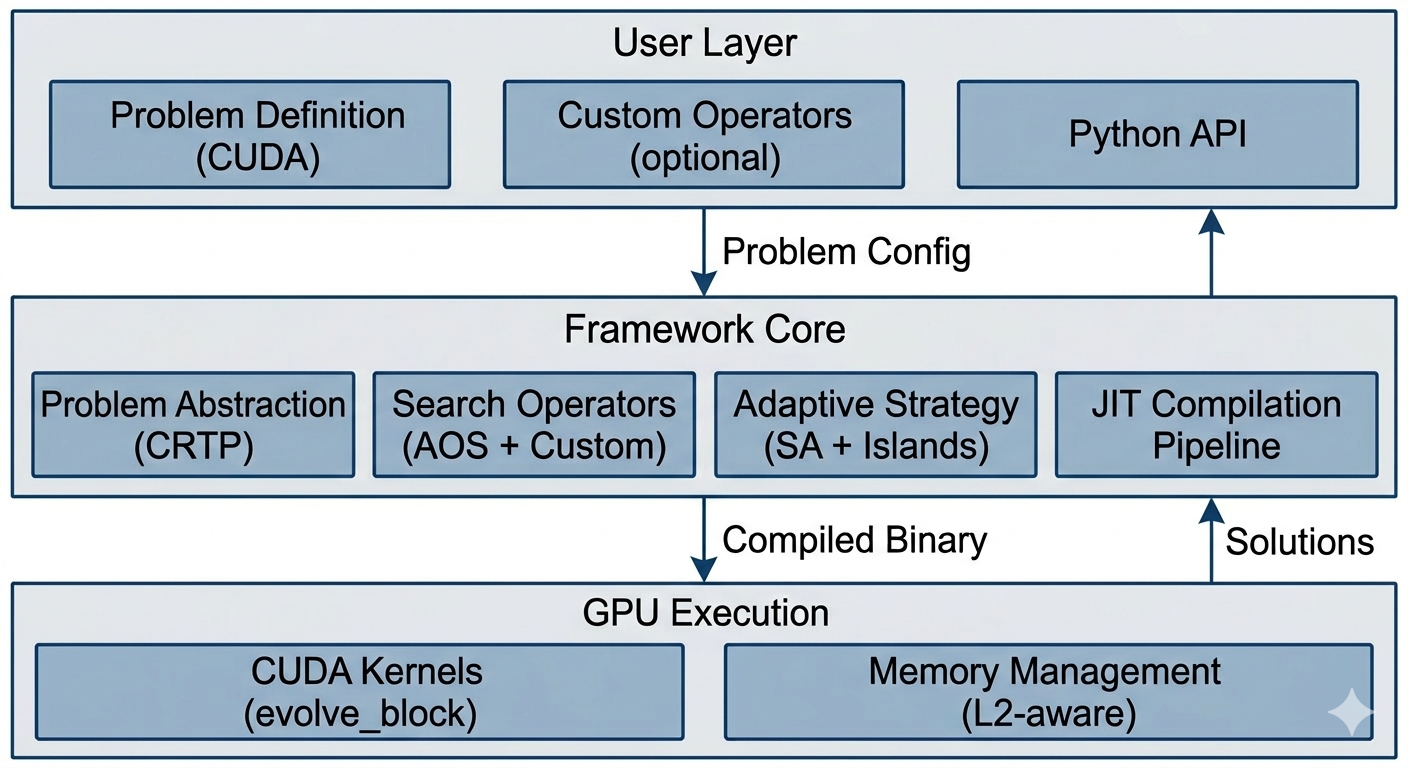}
\caption{cuGenOpt system architecture. The framework provides three-layer abstraction: user interface (Python API and CUDA problem definition), core components (adaptive search and operator management), and GPU execution engine (L2-aware memory management and CUDA kernels).}
\label{fig:system_arch}
\end{figure}

\subsection{Problem Abstraction}
\label{sec:problem_abstraction}

cuGenOpt uses the Curiously Recurring Template Pattern (CRTP)
to define its problem interface.
Users inherit from \texttt{ProblemBase<Derived, D1, D2>}
and implement a small number of functions to define a new problem.

\subsubsection{Encoding Types}

The framework supports three universal encoding types:

\begin{itemize}[leftmargin=*]
  \item \textbf{Permutation}: Solutions are permutations of $[0, n)$,
    suitable for TSP, QAP, Assignment, etc.
  \item \textbf{Binary}: Solutions are $\{0,1\}^n$ vectors,
    suitable for Knapsack, Scheduling, etc.
  \item \textbf{Integer}: Solutions are bounded integer vectors $[lb, ub]^n$,
    suitable for Graph Coloring, Bin Packing, etc.
\end{itemize}

\subsubsection{Solution Structure}

Solutions use a two-dimensional template structure \texttt{Solution<D1, D2>}:

\begin{equation}
\texttt{Solution} = \left\{
\begin{array}{l}
\texttt{data}[D_1][D_2] \quad \text{(solution vectors, row-organized)} \\
\texttt{dim2\_sizes}[D_1] \quad \text{(effective length per row)} \\
\texttt{objectives}[\text{MAX\_OBJ}] \quad \text{(objective values)} \\
\texttt{penalty} \quad \text{(constraint violation penalty)}
\end{array}
\right.
\end{equation}

where $D_1$ is the number of rows (1 for single-sequence problems like TSP,
vehicle count for VRP) and $D_2$ is the maximum row length.
This two-dimensional structure unifies single-sequence (TSP)
and multi-sequence (VRP, JSP) problem representations.

\subsubsection{User Interface}

Users implement only three core functions:
\texttt{compute\_obj(i, sol)} computes the $i$-th objective value,
\texttt{compute\_penalty(sol)} computes constraint violation penalty
(0 indicates feasibility),
and \texttt{config()} returns the problem configuration.
A complete TSP definition requires approximately 20 lines of CUDA (Listing~\ref{lst:tsp}).

\begin{lstlisting}[caption={TSP problem definition (simplified).},label={lst:tsp}]
struct TSPProblem : ProblemBase<TSPProblem, 1, 64> {
  static constexpr ObjDef OBJ_DEFS[] = {
    {"tour_length", ObjDir::Minimize, 1.0f}
  };
  __device__ float compute_obj(int, const Sol& s) const {
    float dist = 0;
    for (int i = 0; i < n-1; i++)
      dist += d_dist[s.data[0][i]*n + s.data[0][i+1]];
    dist += d_dist[s.data[0][n-1]*n + s.data[0][0]];
    return dist;
  }
  __device__ float compute_penalty(const Sol&) const {
    return 0;
  }
};
\end{lstlisting}

\subsection{GPU Parallelization Strategy}
\label{sec:gpu_parallel}

cuGenOpt adopts a \textbf{``one block evolves one solution''} parallel architecture,
balancing population-level and neighborhood-level parallelism.

\subsubsection{Block-Level Architecture}

The $P$ solutions in the population are assigned to $P$ CUDA blocks,
each evolving independently.
Within each block, $T$ threads (default $T{=}128$) independently sample
and evaluate candidate moves,
then perform a block-level reduction to select the best move,
with thread~0 deciding acceptance.

Key advantages:
(1)~the current solution resides in shared memory,
accessible by all threads at $\sim$20-cycle latency
(vs.\ $\sim$400 cycles for global memory);
(2)~$T$ candidate moves per generation provide ample neighborhood sampling,
equivalent to $T\times$ the evaluation volume of serial search;
(3)~blocks are fully independent, requiring no global synchronization.

\begin{figure}[htbp]
\centering
\includegraphics[width=0.95\columnwidth]{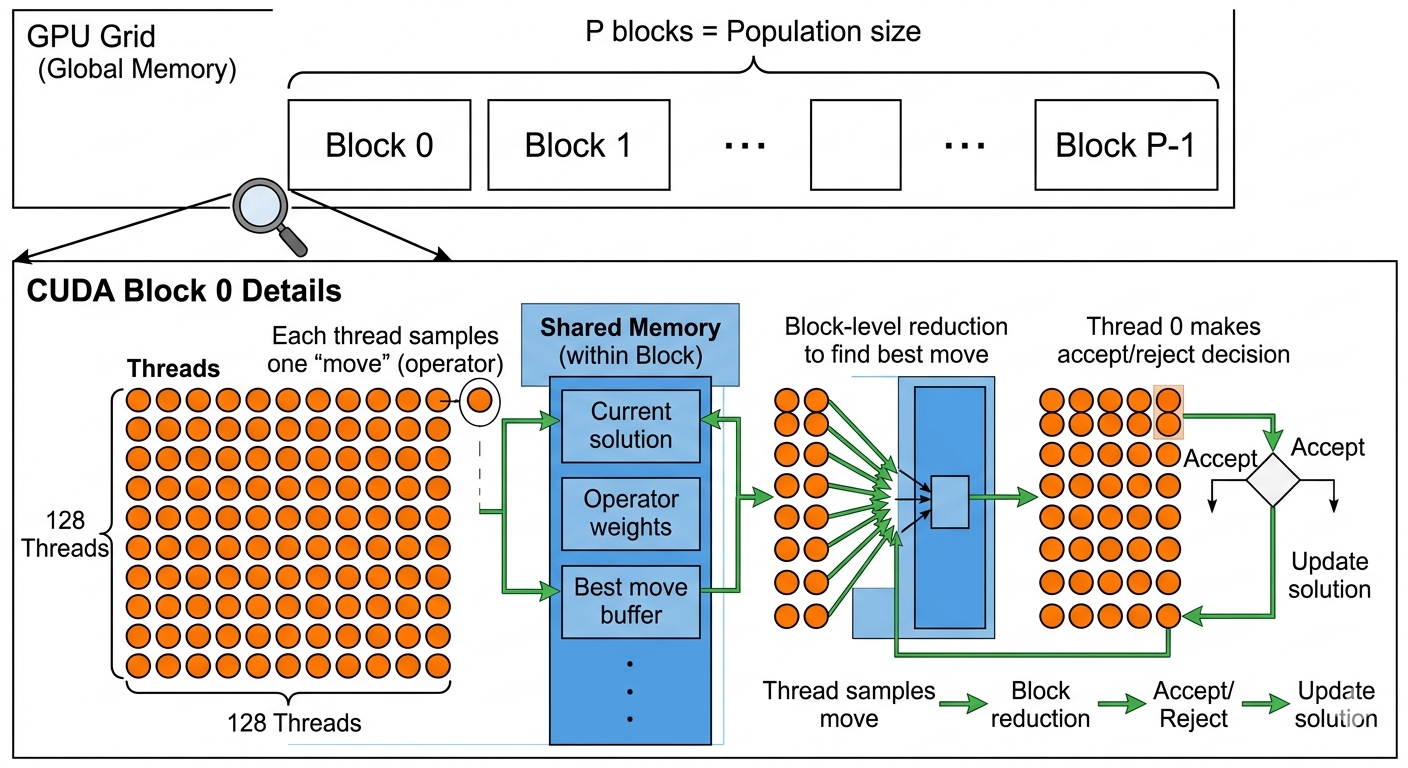}
\caption{GPU parallel execution model. The population is distributed across CUDA blocks (P blocks = population size). Within each block, 128 threads sample candidate moves in parallel, perform block-level reduction to find the best move, and thread 0 executes accept/reject decision. Shared memory holds current solution, operator weights, and reduction buffer.}
\label{fig:gpu_parallel}
\end{figure}

\subsubsection{Shared Memory Layout}

Each block's shared memory is organized as:

\begin{equation}
\underbrace{\texttt{Solution}}_{\text{current sol.}} \;|\;
\underbrace{\texttt{ProblemData}}_{\text{problem data}} \;|\;
\underbrace{\texttt{Candidates}[T]}_{\text{reduction buffer}} \;|\;
\underbrace{\texttt{AOSStats}}_{\text{statistics}}
\end{equation}

When problem data fits in shared memory, all accesses are on-chip.
When data exceeds capacity, the framework automatically falls back
to global memory (via L2 cache),
making the kernel Memory-Bound---the fundamental cause of performance
degradation on large instances.

\subsubsection{Per-Generation Evolution}

Algorithm~\ref{alg:evolve} describes the per-generation evolution within a single block.

\begin{algorithm}[htbp]
\SetAlgoLined
\KwIn{Current solution $s$ in shared memory, sequence registry $\mathcal{R}$,
      K-step weights $\mathbf{w}_K$, temperature $T$}
\KwOut{Updated solution $s$}

\ForEach{thread $t \in \{0, \ldots, T{-}1\}$ \textbf{in parallel}}{
  Sample step count $k_t \in \{1, 2, 3\}$ from $\mathbf{w}_K$\;
  $s_t \leftarrow \text{copy}(s)$ \tcp*{local memory copy}
  \For{$i = 1$ \KwTo $k_t$}{
    Sample sequence $\sigma_i$ from $\mathcal{R}.\mathbf{w}$\;
    Execute $\sigma_i$ on $s_t$\;
  }
  Evaluate $s_t$; compute $\delta_t = f(s_t) - f(s)$\;
}
\textbf{Block reduction}: $t^* = \arg\min_t \delta_t$\;
\If(\tcp*[f]{Thread 0}){$\delta_{t^*} < 0$ \textbf{or} $\text{rand}() < e^{-\delta_{t^*}/T}$}{
  $s \leftarrow s_{t^*}$; update AOS statistics\;
}
$T \leftarrow \alpha \cdot T$\;
\caption{Block-level per-generation evolution (\texttt{evolve\_block\_kernel}).}
\label{alg:evolve}
\end{algorithm}

\subsection{Search Operator System}
\label{sec:operators}

\subsubsection{Built-in Operator Hierarchy}

cuGenOpt organizes search operators into four granularity levels
(Table~\ref{tab:operators}),
with each encoding type having its own operator set.
All operators are identified by a unified \textbf{Sequence ID},
and AOS tracks usage frequency and improvement at the Sequence level.

\begin{table}[htbp]
\centering
\caption{Search operator hierarchy.}
\label{tab:operators}
\small
\begin{tabular}{llp{7cm}}
\toprule
\textbf{Level} & \textbf{Scope} & \textbf{Representative Operators} \\
\midrule
Element & Single element & swap, insert, reverse (Perm); flip (Binary); random\_reset (Int) \\
Segment & Contiguous segment & or-opt, 3-opt (Perm); seg\_flip (Binary); seg\_reset (Int) \\
Row     & Entire row & row\_swap, row\_split, row\_merge (encoding-agnostic) \\
Crossover & Two solutions & OX crossover (Perm); uniform crossover (Binary/Int) \\
\bottomrule
\end{tabular}
\end{table}

The framework also integrates Large Neighborhood Search (LNS)
operators~\cite{pisinger2019large}:
segment shuffle, scatter shuffle, and guided rebuild,
with destruction scope adaptively controlled by problem size.

\subsubsection{User-Defined Operator Registration}
\label{sec:custom_ops}

Built-in operators act in encoding space without exploiting problem semantics.
For performance-sensitive scenarios, domain experts may wish to inject
problem-specific search operators
(e.g., delta-evaluation 2-opt for TSP).
cuGenOpt provides a user-defined operator registration mechanism
that seamlessly integrates such operators into the AOS framework.

\textbf{Registration interface.}
Users encapsulate a CUDA code snippet in a \texttt{CustomOperator} object,
specifying an operator name and Sequence ID ($\geq 100$,
separated from the built-in ID space).
Multiple custom operators can be registered simultaneously.

\textbf{JIT injection.}
Registered operator code is injected into the \texttt{execute\_sequence}
function's switch-case: built-in operators occupy cases $[0, 31]$,
custom operators occupy cases $[\geq 100]$.
Initial weights are registered in \texttt{SeqRegistry}
to participate in AOS EMA weight competition.

\textbf{Problem data access.}
Custom operators need access to problem-specific data (e.g., distance matrices),
but the core framework's operator call chain is problem-agnostic.
cuGenOpt resolves this tension through \textbf{type erasure}:
\texttt{evolve\_block\_kernel} passes the Problem instance's address
as \texttt{const void* prob\_data} to \texttt{execute\_sequence};
in the custom operator's execution function,
\texttt{static\_cast<const CustomProblem*>} recovers the concrete type.
Built-in operators do not use this parameter,
and the compiler optimizes it away.

\textbf{Safety mechanism.}
User-provided CUDA code may contain syntax errors or type mismatches.
On JIT compilation failure, the framework automatically excludes
the offending operator, emits a \texttt{RuntimeWarning},
and falls back to built-in operators only.

\textbf{Isolation design.}
Custom operators are decoupled from the normal path through three layers:
(1)~compile-time \texttt{\#ifdef}---custom operator code is not compiled
when no custom operators are registered;
(2)~runtime defaults---\texttt{prob\_data} defaults to \texttt{nullptr},
and built-in operators never read it;
(3)~Python-layer short-circuit---\texttt{custom\_operators} defaults to \texttt{None},
bypassing all related logic.
Testing confirms zero impact on correctness, performance,
and compilation behavior when no custom operators are present.

\subsection{Population Management}
\label{sec:population}

\subsubsection{Oversample Initialization}

Population initialization uses an oversample-then-select strategy:
$K \times P$ candidate solutions are generated (default $K{=}4$),
evaluated, and the best $P$ are selected.
Selection automatically switches between
single-objective sorting and NSGA-II~\cite{deb2002fast}
non-dominated sorting based on the number of objectives.

\subsubsection{Attribute-Agnostic Heuristic Initialization}
\label{sec:heuristic_init}

Random initialization produces extremely poor initial solutions
on large instances (e.g., TSP-442 random permutation gap $>$ 300\%).
However, as a general-purpose framework, cuGenOpt is unaware of problem semantics
and cannot use problem-specific greedy heuristics.

We propose an \textbf{attribute-agnostic bidirectional construction} method:
for any data matrix $M \in \mathbb{R}^{n \times n}$ provided by the problem,
compute row sums $r_i = \sum_j M_{ij}$ and column sums $c_j = \sum_i M_{ij}$,
then sort in ascending/descending order to generate candidate permutations---four
candidates per matrix, injected into the oversample pool.

\textbf{Intuition.}
Row sums reflect each element's ``average affinity'' with the global structure.
For a TSP distance matrix, cities with small row sums tend to be centrally located;
sorting by row sum clusters geographically proximate cities,
producing solutions far superior to random permutations.
Crucially, this reasoning does not require the framework to know
that the matrix represents distances---for any data matrix,
row/column-sum sorting tends to cluster ``similar'' elements.

This method reduces pcb442 30-second gap from 36\% to 6\%
(Section~\ref{sec:ablation}).

\subsubsection{Island Model and Elite Injection}

The population can be partitioned into islands~\cite{whitley1999island},
each evolving independently with periodic migration of elite solutions.
Three migration strategies are supported: ring, global top-$N$, and hybrid.
Elite injection periodically replaces the worst individual
with the global best solution,
preventing loss of optimal information when simulated
annealing~\cite{kirkpatrick1983optimization} accepts inferior solutions.

\section{Adaptive Search and Hardware Awareness}
\label{sec:adaptive}

A core challenge for general-purpose solvers is that optimal search strategies
vary significantly across problem types and scales.
This section presents cuGenOpt's adaptive search mechanisms
and hardware-aware resource management,
which jointly optimize search strategy and GPU utilization.

\subsection{Problem-Profile-Driven Prior Strategy (L1)}
\label{sec:problem_profile}

Different problem scales tolerate different operator complexities.
For example, 3-opt has $O(n^3)$ time complexity---negligible at $n{=}50$
but severely throttling search speed at $n{=}400$.
Starting all operators with equal weights forces AOS
to spend many iterations learning this prior.

cuGenOpt statically analyzes the \texttt{ProblemConfig} structure
to classify each problem into a \textbf{problem profile}:

\begin{definition}[Problem Profile]
$\mathcal{P} = (\text{encoding}, \text{scale}, \text{structure}, p_{\text{cross}})$, where
$\text{scale} \in \{\text{Small}, \text{Medium}, \text{Large}\}$
is determined by $D_2$ ($\leq 100$, $\leq 250$, $> 250$),
and $\text{structure} \in \{\text{SingleSeq}, \text{MultiFixed}, \text{MultiPartition}\}$
is determined by row count and row mode.
\end{definition}

Each scale class maps to a weight preset (Table~\ref{tab:weight_preset})
controlling initial weights for operators of different complexities.
$O(1)$ operators (swap, insert, etc.) are unaffected by scale;
high-complexity operators (3-opt, LNS) are heavily down-weighted at large scale.
This ensures lightweight operators receive more sampling budget on large problems,
while not completely disabling complex operators---AOS can still
up-weight them at runtime based on observed effectiveness.

\begin{table}[htbp]
\centering
\caption{Scale-driven operator weight presets.}
\label{tab:weight_preset}
\small
\begin{tabular}{lcccc}
\toprule
\textbf{Scale} & \textbf{3-opt} ($O(n^3)$) & \textbf{or-opt} ($O(n^2)$) & \textbf{LNS} & \textbf{LNS cap} \\
\midrule
Small ($D_2 \leq 100$)        & 0.50 & 0.80 & 0.006 & 0.02 \\
Medium ($100 < D_2 \leq 250$) & 0.30 & 0.70 & 0.004 & 0.01 \\
Large ($D_2 > 250$)           & 0.05 & 0.30 & 0.001 & 0.005 \\
\bottomrule
\end{tabular}
\end{table}

\subsection{Adaptive Operator Selection (L3 AOS)}
\label{sec:aos}

\subsubsection{Two-Level Weight Structure}

cuGenOpt's AOS maintains two levels of weights:
(1)~\textbf{K-step level}: weights $\mathbf{w}_K = (w_1, w_2, w_3)$
control the number of operators executed per iteration---$K{=}1$
for single-step search, $K{=}2,3$ for multi-step composite search;
(2)~\textbf{Sequence level}: weights $\mathbf{w}_S$ control
the selection probability of each specific operator,
with per-sequence weight caps $\bar{w}_i$.

Custom operators (Section~\ref{sec:custom_ops}) automatically receive
Sequence-level weights upon registration,
competing within the same AOS framework as built-in operators.

\begin{figure}[htbp]
\centering
\includegraphics[width=0.95\columnwidth]{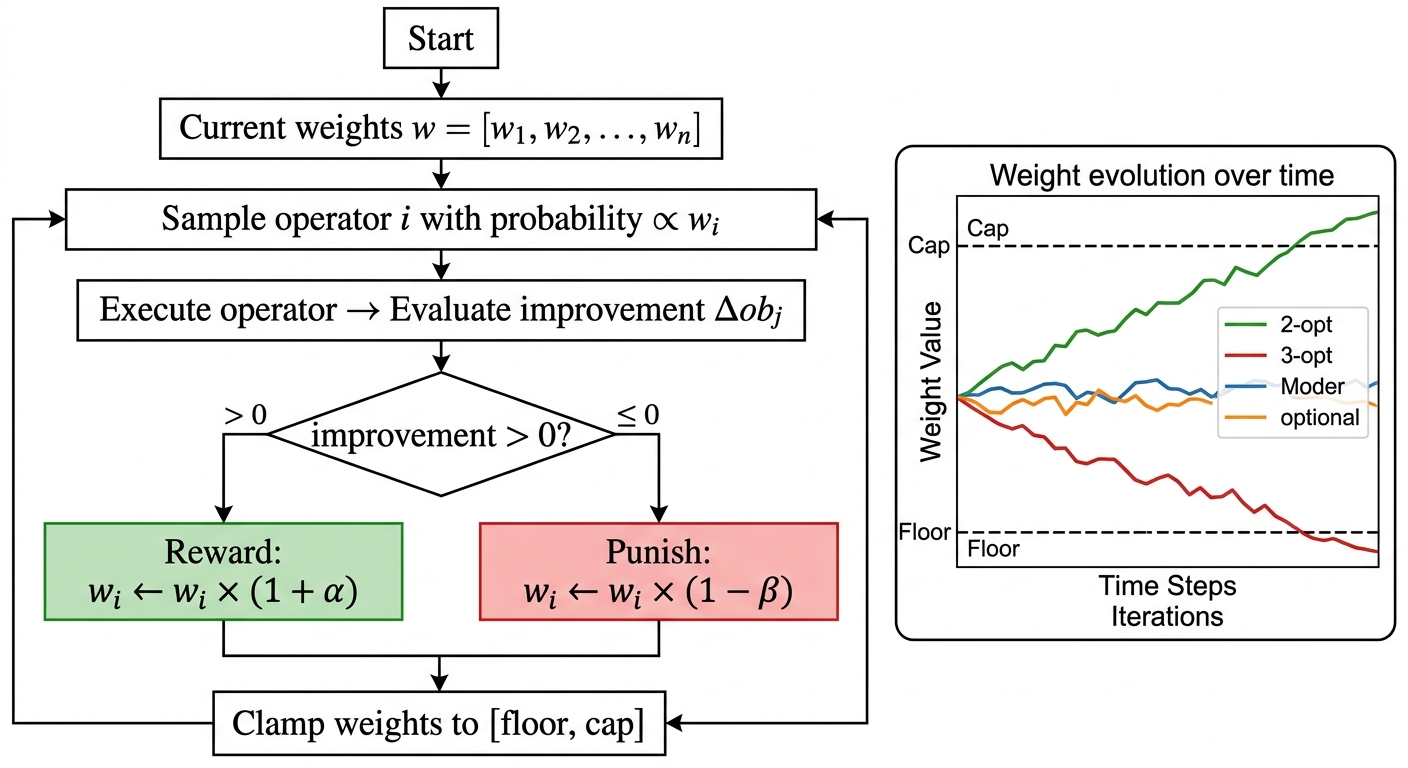}
\caption{Adaptive Operator Selection (AOS) mechanism. Each CUDA block maintains local statistics (usage count and improvement count) for each operator. Every 10 batches, weights are updated via EMA based on observed effectiveness. The two-level structure controls both K-step (number of operators per iteration) and sequence-level (specific operator) selection probabilities.}
\label{fig:aos_mechanism}
\end{figure}

\subsubsection{Statistics Collection and Weight Update}

Each CUDA block maintains local AOS statistics in shared memory:
usage count $u_i$ and improvement count $v_i$ per Sequence.
Collection is performed entirely on the GPU via atomic operations,
requiring no CPU--GPU communication.

Every $I$ batches (default $I{=}10$), weights are updated via EMA:

\begin{equation}
w_i^{(t+1)} = \alpha \cdot w_i^{(t)} + (1 - \alpha) \cdot \left(\frac{v_i}{u_i + \epsilon} + w_{\text{floor}}\right)
\label{eq:ema}
\end{equation}

Updated weights are clamped to $[w_{\text{floor}}, \min(w_{\text{cap}}, \bar{w}_i)]$
and normalized.
The choice of update interval $I$ is critical:
$I{=}1$ introduces frequent \texttt{cudaMemcpy} synchronization overhead;
overly large $I$ makes weights unresponsive.
Experiments show $I{=}10$ achieves the best balance
(Section~\ref{sec:ablation}).

\subsubsection{Stagnation Detection}

When multiple consecutive batches yield no improvement,
K-step weights are reset to defaults
($w_1{=}0.8, w_2{=}0.15, w_3{=}0.05$),
increasing multi-step search probability to escape local optima.

Algorithm~\ref{alg:aos} summarizes the complete AOS procedure.

\begin{algorithm}[htbp]
\SetAlgoLined
\KwIn{Sequence registry $\mathcal{R}$, K-step config $\mathcal{K}$, update interval $I$}
\For{each batch $b$}{
  Run \texttt{evolve\_block\_kernel}; blocks accumulate AOS stats in shared memory\;
  \If{$b \bmod I = 0$}{
    Aggregate statistics from all blocks;
    update $\mathcal{R}.\mathbf{w}$ and $\mathcal{K}.\mathbf{w}$ via Eq.~\eqref{eq:ema}\;
    Clamp and normalize weights\;
    \If{consecutive no-improvement $>$ threshold}{
      Reset $\mathcal{K}.\mathbf{w}$ to defaults\;
    }
    Write updated parameters back to device\;
  }
}
\caption{Adaptive Operator Selection (AOS) procedure.}
\label{alg:aos}
\end{algorithm}

\subsection{Shared Memory Auto-Extension}
\label{sec:smem_extension}

CUDA defaults to 48\,KB of shared memory per block.
When problem data exceeds this limit, the standard approach
is to fall back to global memory,
incurring approximately $20\times$ latency penalty.
However, modern GPUs support configurable shared memory
well beyond 48\,KB (Table~\ref{tab:gpu_smem}),
accessible via \texttt{cudaFuncSetAttribute}.

\begin{table}[htbp]
\centering
\caption{Maximum shared memory per block across GPU generations.}
\label{tab:gpu_smem}
\small
\begin{tabular}{lrrl}
\toprule
\textbf{GPU} & \textbf{SMs} & \textbf{Max Shared Mem/Block} & \textbf{Architecture} \\
\midrule
Tesla T4     & 40  & 64\,KB  & Turing (sm\_75) \\
Tesla V100   & 80  & 96\,KB  & Volta (sm\_70) \\
A800/A100    & 108 & 164\,KB & Ampere (sm\_80) \\
H20/H100     & 132 & 228\,KB & Hopper (sm\_90) \\
\bottomrule
\end{tabular}
\end{table}

cuGenOpt implements a three-layer separation of concerns:
the Problem reports ``I need $X$ bytes of shared memory'' without regard to hardware;
the Solver attempts to request $X$ bytes from CUDA (\texttt{cudaFuncSetAttribute});
the Solver then decides on overflow based on the actual allocation result---falling
back to global memory only if the request fails.

This mechanism is fully transparent to users.
Experiments show it boosts VRPTW throughput by 75--81\% on T4
(Section~\ref{sec:ablation}).

\subsection{Adaptive Population Sizing}
\label{sec:pop_sizing}

Population size $P$ determines the number of concurrent solutions,
directly affecting search diversity and per-generation throughput.
cuGenOpt automatically computes $P$ based on a two-level memory hierarchy:

\textbf{Shared memory path.}
When problem data fits in shared memory,
$P$ is determined by \texttt{cudaOccupancyMaxActiveBlocksPerMultiprocessor}
to maximize GPU occupancy.

\textbf{Global memory path.}
When data overflows shared memory,
each block's working set competes for L2 cache bandwidth.
The key insight is that L2 cache capacity relative to working set size
determines the effective population ceiling:

\begin{equation}
P = \begin{cases}
P_{\text{SM}} & \text{if} \; L2_{\text{size}} / W \geq P_{\text{SM}} / 2 \\[4pt]
\lfloor L2_{\text{size}} / W \rfloor_{\text{pow2}} & \text{otherwise}
\end{cases}
\label{eq:pop_sizing}
\end{equation}

\noindent
where $P_{\text{SM}} = 2^{\lceil \log_2(\text{SM count}) \rceil}$
and $W$ is the per-block working set size.
The condition ensures that SM-derived population size is used
only when L2 cache can sustain at least half the concurrent blocks;
otherwise, population is reduced to prevent cache thrashing.

Without this logic, V100 (80 SMs, $P_{\text{SM}}{=}128$)
achieves only 368~gens/s and 57.88\% gap on pcb442 (working set 763\,KB),
while $P{=}32$ achieves 811~gens/s and 5.29\% gap (Section~\ref{sec:ablation}).

\begin{figure}[htbp]
\centering
\includegraphics[width=0.95\columnwidth]{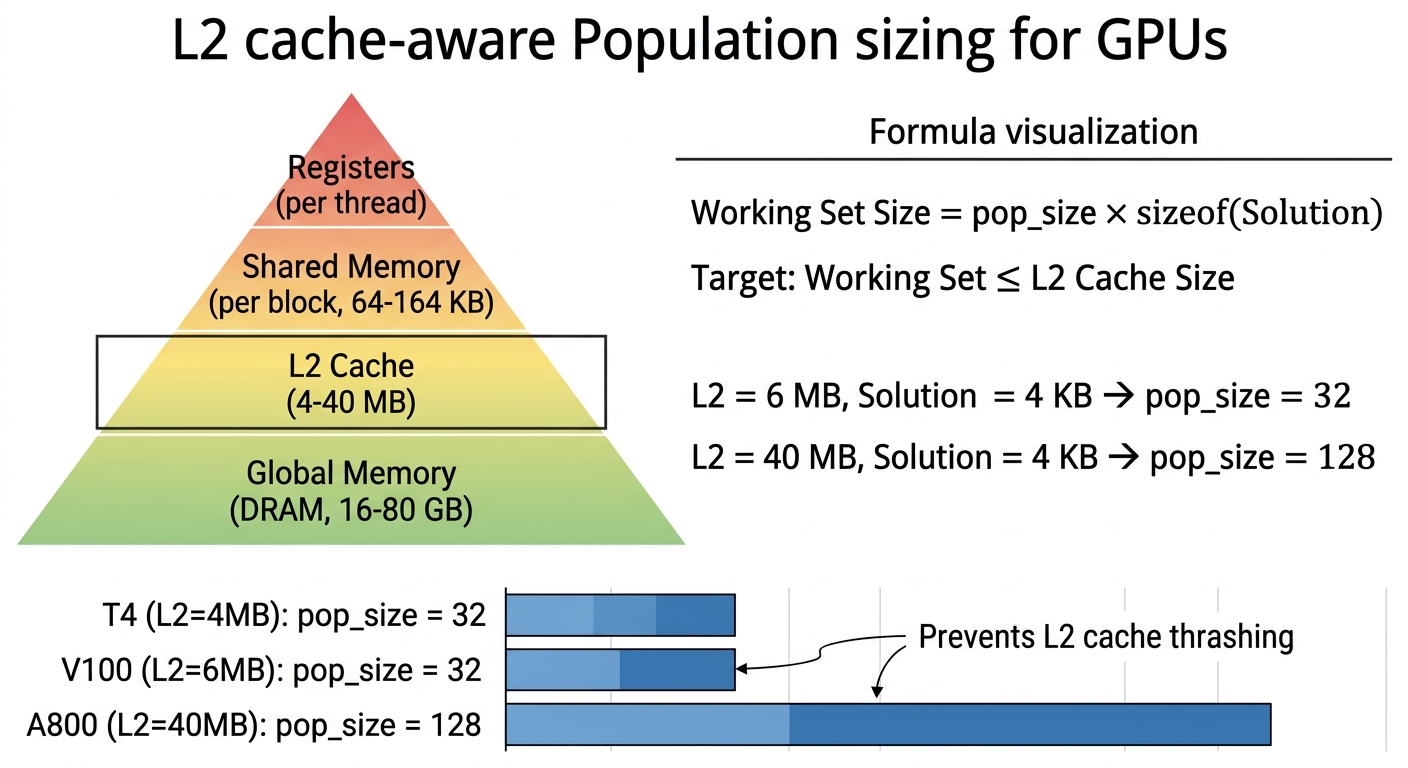}
\caption{L2 cache-aware population sizing. The framework automatically adjusts population size to fit the working set into L2 cache, preventing cache thrashing. For TSP n=1000 (working set 3.8 MB), V100 (L2=6 MB) selects pop=32, while A800 (L2=40 MB) can support larger populations.}
\label{fig:l2_aware}
\end{figure}

\section{User Interface and Toolchain}
\label{sec:interface}

The cuGenOpt core engine is implemented as a header-only CUDA C++ library.
To lower the barrier to adoption, this section describes three progressive
interface layers: a JIT compilation pipeline (wrapping CUDA snippets as Python calls),
a Python API (one-line solving for built-in problems),
and an LLM-based modeling assistant (natural language to GPU solving).

\subsection{JIT Compilation Pipeline}
\label{sec:jit}

cuGenOpt's Python package is built on \textbf{Just-In-Time (JIT) compilation}
rather than traditional pybind11 pre-compiled bindings.
This design choice stems from a fundamental property of the framework:
user-defined objective functions and constraints are CUDA code snippets
that must be compiled together with the framework code
to achieve full GPU acceleration.

\textbf{Pipeline.}
The Python layer fills user-provided parameters
(encoding type, dimensions, objective function code, data arrays, etc.)
into a predefined \texttt{.cu} template,
invokes \texttt{nvcc} to compile a standalone executable,
runs it via \texttt{subprocess}, and parses JSON-formatted output.
The entire process is transparent to the user---the API call
behaves like an ordinary Python function.

\textbf{Compilation cache.}
A SHA-256 hash of the generated \texttt{.cu} source ensures
that identical problem definitions are not recompiled.
First compilation takes approximately 9 seconds (including \texttt{nvcc} startup);
cache hits reduce this to approximately 0.1 seconds.

\textbf{Custom operator integration.}
When users register custom operators (Section~\ref{sec:custom_ops}),
the JIT pipeline injects operator code into the template's
\texttt{execute\_custom\_op} function,
defines the \texttt{CUGENOPT\_HAS\_CUSTOM\_OPS} macro,
and injects initial weights into the \texttt{register\_custom\_operators} callback.
Compilation failures trigger automatic exclusion and fallback.

\textbf{Design rationale.}
Compared to pre-compiled bindings, the JIT pipeline offers three advantages:
(1)~\textbf{cross-platform}---a pure-Python package (\texttt{py3-none-any}),
with identical workflows on Linux, Windows, and macOS,
and no platform-specific binaries;
(2)~\textbf{minimal footprint}---the wheel is approximately 83\,KB
(including framework headers),
far smaller than pre-compiled \texttt{.so} files;
(3)~\textbf{full performance}---user code is compiled together with
the framework, achieving runtime performance equivalent to hand-written CUDA
with no Python--C++ call overhead.

\begin{figure}[htbp]
\centering
\includegraphics[width=0.95\columnwidth]{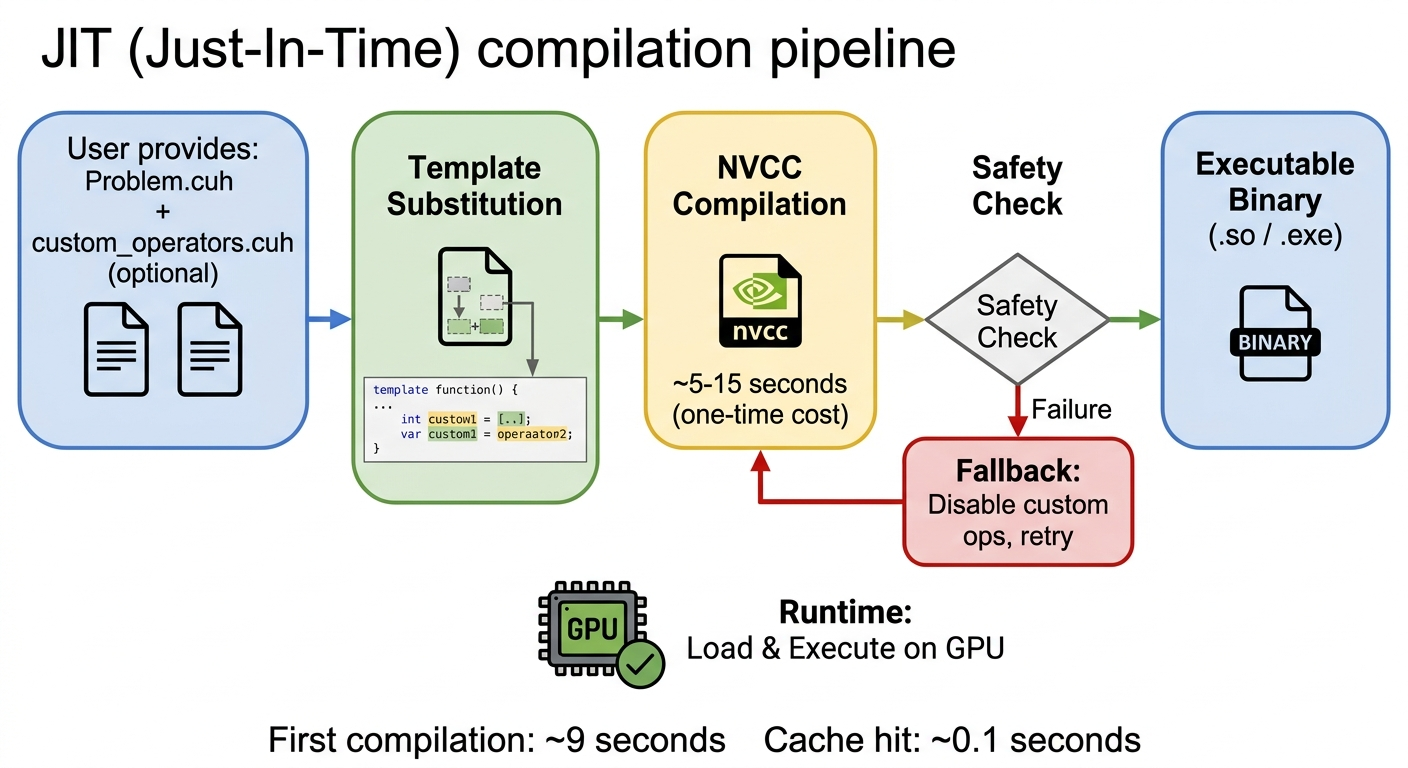}
\caption{JIT compilation pipeline. User-provided CUDA code snippets (objective and constraints) are injected into a template, compiled via nvcc, and executed as a subprocess. SHA-256 hashing enables compilation caching. The entire process is transparent to users, appearing as a standard Python function call.}
\label{fig:jit_pipeline}
\end{figure}

\subsection{Python API}
\label{sec:python_api}

Building on the JIT pipeline, cuGenOpt provides two levels of Python API:

\textbf{Built-in problems.}
Nine standard problems (TSP, VRP, VRPTW, Knapsack, QAP, etc.)
have pre-written CUDA code snippets wrapped as \texttt{solve\_xxx} functions:

\begin{lstlisting}[language=Python,caption={Python API examples.},label={lst:python_api}]
import cugenopt
result = cugenopt.solve_tsp(dist_matrix, time_limit=30)
result = cugenopt.solve_knapsack(weights, values, cap)
\end{lstlisting}

\textbf{Custom problems.}
Users provide CUDA code snippets defining objectives and constraints:

\begin{lstlisting}[language=Python,caption={Custom problem example.},label={lst:custom}]
result = cugenopt.solve_custom(
    encoding="permutation", dim2=64, n=50,
    compute_obj="...",      # CUDA code snippet
    compute_penalty="...",  # CUDA code snippet
    data={"d_dist": distance_matrix},
    custom_operators=[...], # optional
    time_limit=30,
)
\end{lstlisting}

\subsection{LLM-Based Modeling Assistant}
\label{sec:ai_modeling}

While the Python API substantially lowers the barrier,
users still need to write CUDA code snippets for objective functions.
To eliminate this requirement entirely,
we develop an \textbf{LLM-based modeling assistant}
that converts natural-language problem descriptions
into compilable and executable cuGenOpt solver code.

The tool is implemented as a structured instruction set (Agent Skill)
that guides an LLM programming agent through a three-phase workflow:
(1)~\textbf{Requirement analysis}---extract decision variables,
objectives, and constraints from natural language;
select encoding type and dimensions.
(2)~\textbf{Code generation}---produce problem definition code;
for moderately complex problems,
summarize the generated logic in natural language for user confirmation.
(3)~\textbf{Validation and execution}---automatically detect
the GPU environment (local or SSH remote),
compile, run, and return results.

The tool infers \emph{execution depth} from user phrasing:
``generate'' produces code only,
while ``run it'' triggers the full compile--execute pipeline.
In validation testing, the input
``I have a 20-city TSP problem, run it for me''
produces an end-to-end feasible solution
(tour length 410.16, 177\,ms) with zero user-written code.

This demonstrates that when an LLM agent is equipped
with structured domain knowledge
(encoding taxonomy, API specifications, reference implementations),
it can effectively bridge the gap between domain experts
and GPU-accelerated optimization frameworks.

\section{Experiments}
\label{sec:experiments}

We evaluate cuGenOpt through six thematic experiment suites:
(1)~baseline comparisons,
(2)~scalability and hardware adaptation,
(3)~large-scale and multi-GPU validation,
(4)~generality and standard benchmarks,
(5)~framework optimization ablation,
and (6)~user-defined operator effectiveness.

\subsection{Experimental Setup}
\label{sec:exp_setup}

\subsubsection{Hardware}

\begin{table}[htbp]
\centering
\caption{Experimental environment.}
\label{tab:env}
\small
\begin{tabular}{ll}
\toprule
\textbf{Component} & \textbf{Specification} \\
\midrule
\multicolumn{2}{l}{\textit{GPU platforms (cuGenOpt)}} \\
Tesla T4   & 40 SMs, 64\,KB shared/block, 4\,MB L2, 320\,GB/s \\
V100-SXM2  & 80 SMs, 96\,KB shared/block, 6\,MB L2, 900\,GB/s \\
V100S-PCIE & 80 SMs, 96\,KB shared/block, 6\,MB L2, 1134\,GB/s \\
A800-SXM4  & 108 SMs, 164\,KB shared/block, 40\,MB L2, 2039\,GB/s \\
\midrule
\multicolumn{2}{l}{\textit{CPU platform (baselines)}} \\
CPU & Apple M-series (ARM, high single-thread IPC) \\
MIP solvers & SCIP 8.x~\cite{scip}, CBC (via OR-Tools) \\
Routing solver & OR-Tools Routing~\cite{ortools} (GLS) \\
\midrule
\multicolumn{2}{l}{\textit{Common settings}} \\
Random seeds & 42, 123, 456, 789, 2024 (5 seeds per experiment) \\
\bottomrule
\end{tabular}
\end{table}

\subsubsection{Benchmark Instances}

TSP instances are from TSPLIB~\cite{reinelt1991tsplib} (eil51 through pcb442),
VRP from Augerat~\cite{augerat1995computational},
VRPTW from Solomon~\cite{solomon},
QAP from QAPLIB~\cite{qaplib},
JSP from OR-Library~\cite{orlib},
and Knapsack from Pisinger~\cite{pisinger}.

\subsection{Baseline Comparisons}
\label{sec:baseline}

\subsubsection{vs.\ General MIP Solvers}

Table~\ref{tab:vs_mip} compares cuGenOpt with MIP solvers under a 60-second time limit.

\begin{table}[htbp]
\centering
\caption{cuGenOpt vs.\ MIP solvers (gap\%, 60\,s, Tesla T4).}
\label{tab:vs_mip}
\small
\begin{tabular}{lrrrl}
\toprule
\textbf{Instance} & \textbf{cuGenOpt} & \textbf{SCIP} & \textbf{CBC} & \textbf{Winner} \\
\midrule
eil51 ($n{=}51$)     & \textbf{0.00\%} & 69.5\%  & 31.2\%     & cuGenOpt \\
ch150 ($n{=}150$)    & \textbf{0.34\%} & 709.9\% & infeasible & cuGenOpt \\
A-n32-k5 ($n{=}32$)  & \textbf{0.00\%} & 56.8\%  & ---        & cuGenOpt \\
\bottomrule
\end{tabular}
\end{table}

cuGenOpt dominates across all instances.
The MIP bottleneck is the $O(n^2)$ variable count of the MTZ formulation:
at $n{=}150$, SCIP's 60-second gap remains 710\%, and CBC finds no feasible solution.
We note that stronger MIP formulations or commercial solvers may perform better,
but MTZ represents ``general-purpose MIP modeling''---the kind a user
without algorithm expertise would write---forming a fair comparison
with cuGenOpt's low-barrier positioning.

\subsubsection{vs.\ Specialized Solvers}

Table~\ref{tab:vs_ortools} compares cuGenOpt with OR-Tools Routing.

\begin{table}[htbp]
\centering
\caption{cuGenOpt vs.\ OR-Tools Routing (gap\%, 60\,s, Tesla T4).}
\label{tab:vs_ortools}
\small
\begin{tabular}{lrrrr}
\toprule
\textbf{Instance} & $n$ & \textbf{cuGenOpt} & \textbf{OR-Tools} & \textbf{Winner} \\
\midrule
eil51    & 51  & \textbf{0.00\%} & 0.00\%  & Tie \\
kroA100  & 100 & \textbf{0.00\%} & 0.00\%  & Tie \\
ch150    & 150 & \textbf{0.34\%} & 0.54\%  & cuGenOpt \\
tsp225   & 225 & 2.30\%          & \textbf{$-$0.61\%} & OR-Tools \\
lin318   & 318 & 5.16\%          & \textbf{2.85\%} & OR-Tools \\
pcb442   & 442 & 4.75\%          & \textbf{1.87\%} & OR-Tools \\
A-n32-k5 & 32  & \textbf{0.00\%} & 0.00\%  & Tie \\
\bottomrule
\end{tabular}
\end{table}

\textbf{Key findings.}
(1)~At small scale ($n \leq 100$), cuGenOpt matches the specialized solver.
(2)~At $n{=}150$, cuGenOpt unexpectedly wins (0.34\% vs.\ 0.54\%),
suggesting that GPU parallel search diversity can compensate
for algorithmic specialization on certain medium-scale instances.
(3)~At large scale ($n > 200$), OR-Tools leads,
with the gap primarily attributable to initial solution quality
and search efficiency.

\begin{figure}[htbp]
\centering
\includegraphics[width=0.95\columnwidth]{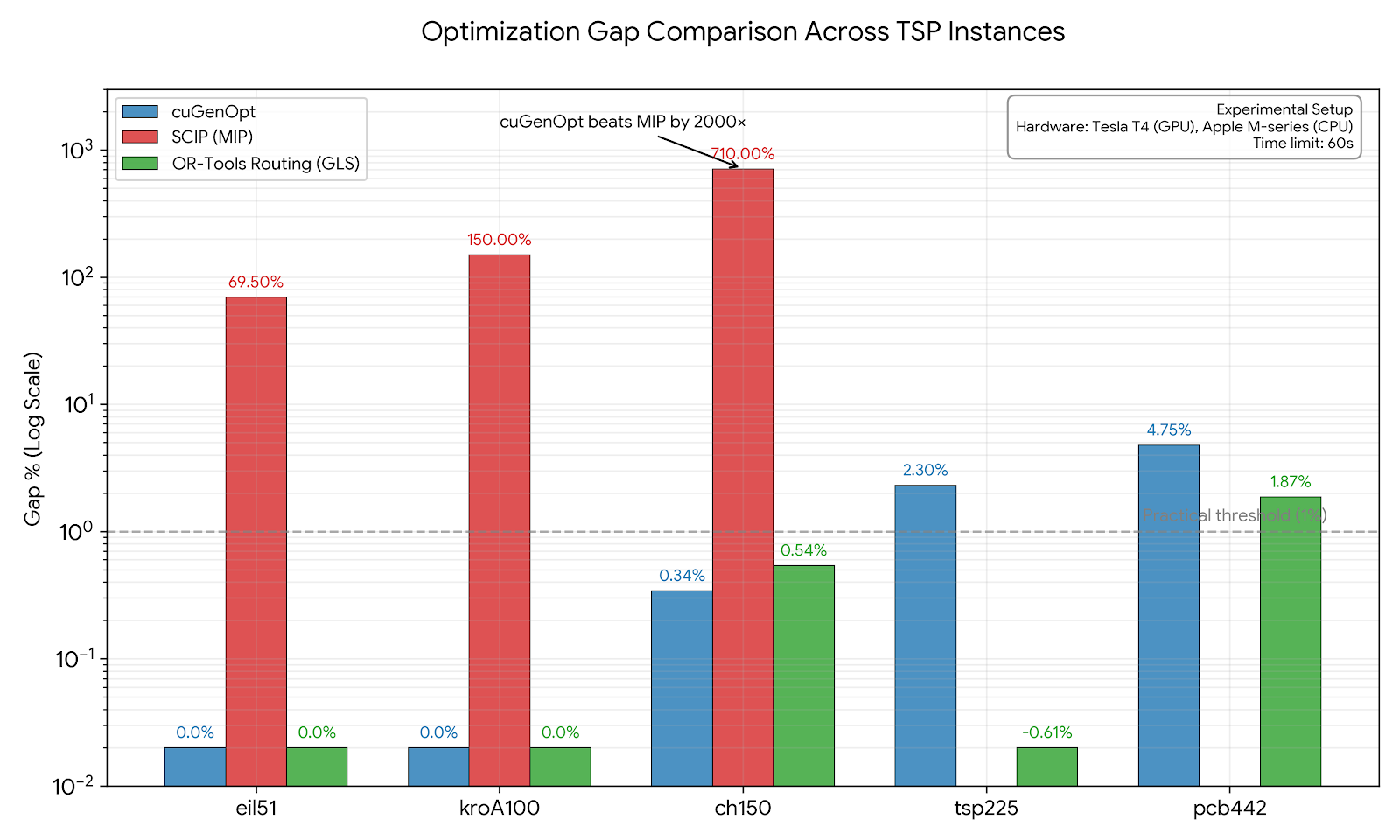}
\caption{Baseline comparison across TSP instances. cuGenOpt dominates general MIP solvers (SCIP/CBC) across all scales, with gaps 100-2000× smaller. Against specialized routing solvers (OR-Tools Routing with GLS metaheuristic), cuGenOpt is competitive at small-medium scale ($n \leq 150$) but trails at large scale ($n > 200$).}
\label{fig:baseline_comparison}
\end{figure}

\subsubsection{Custom Scenarios---Modeling Boundaries of Specialized Solvers}

To demonstrate generality advantages,
we design two scenarios that OR-Tools cannot model precisely:

\textbf{Scenario A: Priority-constrained VRP}---within each vehicle route,
high-priority customers must be visited before low-priority ones.
OR-Tools' Dimension mechanism cannot express intra-route partial orders.
cuGenOpt models this by adding $\sim$10 lines to \texttt{compute\_penalty}.

\textbf{Scenario B: Nonlinear transport cost VRP}---edge cost
grows nonlinearly with cumulative load:
$\text{cost} = d_{ij} \times (1 + 0.3 \times (\text{load}/\text{cap})^2)$.
OR-Tools' \texttt{ArcCostEvaluator} cannot access current vehicle load.
cuGenOpt handles this with $\sim$5 lines in \texttt{compute\_obj}.

\begin{table}[htbp]
\centering
\caption{Custom scenario comparison (A-n32-k5, 60\,s, Tesla T4).}
\label{tab:custom_scenarios}
\small
\begin{tabular}{lcc}
\toprule
\textbf{Metric} & \textbf{cuGenOpt} & \textbf{OR-Tools} \\
\midrule
\multicolumn{3}{l}{\textit{Scenario A: Priority constraints}} \\
Priority violations & \textbf{0} & 17 \\
\midrule
\multicolumn{3}{l}{\textit{Scenario B: Nonlinear cost}} \\
Nonlinear cost & \textbf{826.67} & 874.30 \\
\bottomrule
\end{tabular}
\end{table}

Specialized solvers excel on standard scenarios,
but custom constraints and objectives exceed their modeling capabilities.
cuGenOpt supports these with +10 lines (constraints) or +5 lines (objectives)---a
core advantage of the general-purpose approach.

\subsection{Scalability and Hardware Adaptation}
\label{sec:scalability}

\subsubsection{TSP Scale Expansion}

Table~\ref{tab:scale_gap} presents solution quality for TSP instances
from $n{=}51$ to $442$ across three GPUs.

\begin{table}[htbp]
\centering
\caption{Three-GPU TSP comparison (mean gap\%, 30\,s, 5 seeds).}
\label{tab:scale_gap}
\small
\begin{tabular}{lrrrrrrr}
\toprule
\textbf{Instance} & $n$ & \textbf{T4} & \textbf{V100} & \textbf{A800} & \textbf{T4 g/s} & \textbf{V100 g/s} & \textbf{A800 g/s} \\
\midrule
eil51   & 51  & 0.09 & \textbf{0.00} & \textbf{0.00} & 4{,}160 & 3{,}307 & \textbf{4{,}821} \\
kroA100 & 100 & 0.14 & 0.05 & \textbf{0.02} & \textbf{5{,}717} & 2{,}361 & 3{,}115 \\
ch150   & 150 & 2.02 & \textbf{0.78} & 1.14 & 1{,}698 & 1{,}613 & \textbf{2{,}135} \\
tsp225  & 225 & \textbf{2.82} & \textbf{2.82} & 3.50 & 1{,}781 & \textbf{3{,}102} & 2{,}344 \\
lin318  & 318 & 3.61 & \textbf{3.51} & 6.49 & 996 & \textbf{1{,}470} & 1{,}149 \\
pcb442  & 442 & 6.15 & 5.29 & \textbf{4.73} & 621 & 805 & \textbf{1{,}348} \\
\bottomrule
\end{tabular}
\end{table}

\subsubsection{Memory Hierarchy Analysis}

Cross-hardware comparison reveals three key findings:

\textbf{Shared memory capacity determines the performance boundary.}
ch150 (distance matrix $\sim$87\,KB) exceeds T4's 64\,KB and V100's 96\,KB limits;
both use global memory.
A800's 164\,KB accommodates this data, enabling the shared memory path
with 32\% higher throughput than V100.

\textbf{L2 cache and bandwidth dominate large-instance performance.}
On pcb442 (working set 763\,KB), all three GPUs use global memory.
A800's 40\,MB L2 and 2\,TB/s bandwidth achieve 1{,}348~gens/s
(2.17$\times$ T4), with the best gap of 4.73\%.

\textbf{Population--generation trade-off.}
A800 selects $P{=}128$ on tsp225/lin318 (ample L2),
but lower gens/s yields insufficient total generations,
resulting in worse gap than $P{=}32$ on V100/T4.
L2 capacity alone is insufficient for population sizing---convergence
requirements must also be considered.

\begin{figure}[htbp]
\centering
\includegraphics[width=0.95\columnwidth]{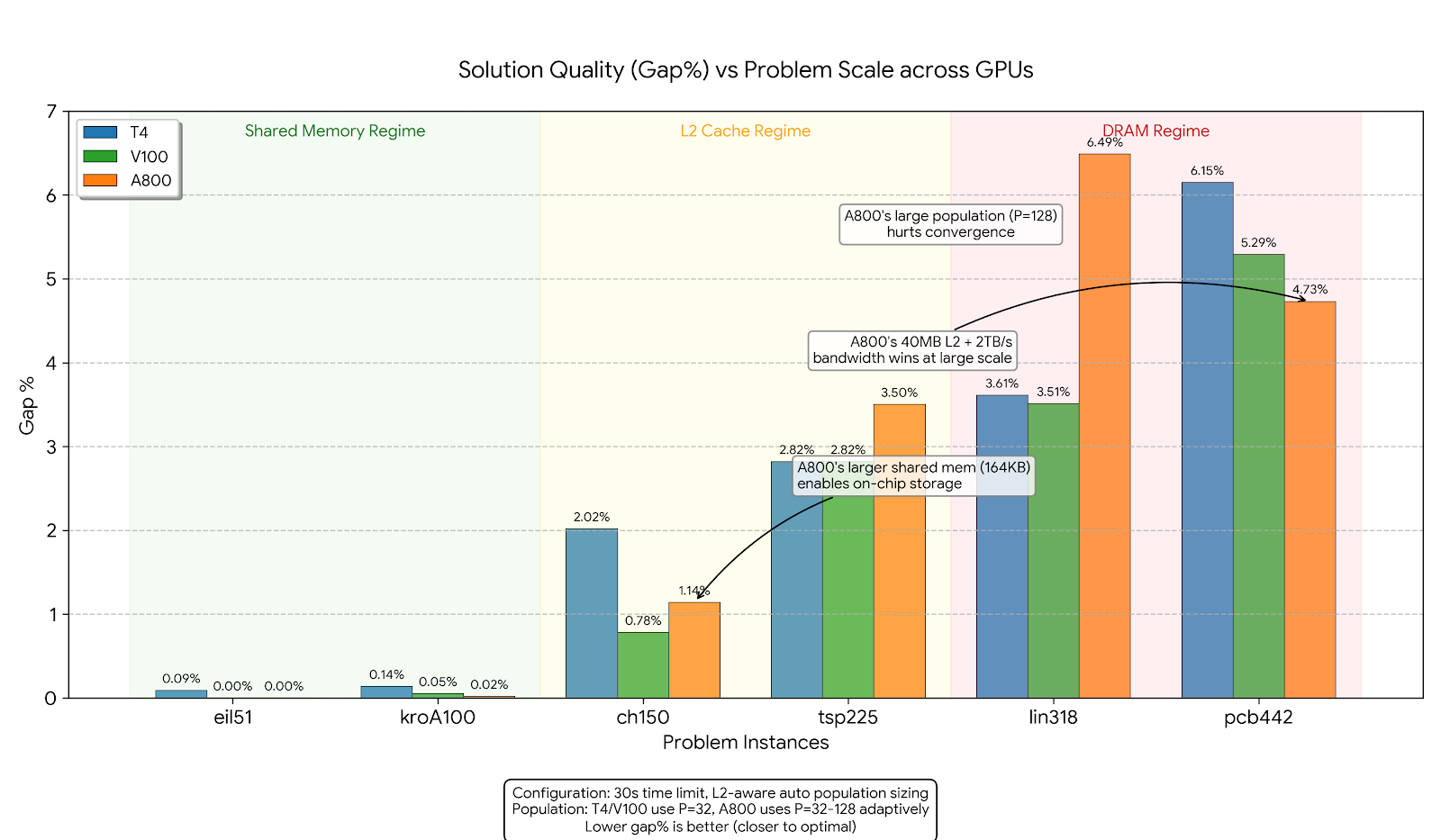}
\caption{Cross-hardware solution quality comparison (30s, gap\%). Three memory regimes emerge: (1) Shared memory regime ($n \leq 100$): A800's 164KB shared memory enables best quality. (2) L2 cache regime (n=150-225): A800's adaptive large population (P=128) hurts convergence despite ample L2. (3) DRAM regime ($n \geq 318$): A800's 40MB L2 and 2TB/s bandwidth achieve best quality at large scale.}
\label{fig:cross_hardware}
\end{figure}

\subsection{Large-Scale and Multi-GPU Validation}
\label{sec:large_scale}

To validate framework scalability and multi-GPU effectiveness,
we conduct experiments on problems up to $n{=}1000$ using 2×V100S GPUs.

\subsubsection{Large-Scale Single-GPU Performance}

Table~\ref{tab:large_scale} presents results for TSP and VRP at scales
beyond standard benchmarks.

\begin{table}[htbp]
\centering
\caption{Large-scale problem results (V100S, 5000 generations).}
\label{tab:large_scale}
\small
\begin{tabular}{lrrrr}
\toprule
\textbf{Problem} & $n$ & \textbf{Solution} & \textbf{Time (s)} & \textbf{Gens/s} \\
\midrule
TSP  & 300  & 8{,}156.07  & 4.2  & 1{,}190 \\
TSP  & 1000 & 77{,}691.52 & 15.3 & 327 \\
VRP  & 150  & 4{,}504.37  & 3.8  & 1{,}316 \\
VRP  & 500  & 56{,}732.55 & 10.1 & 495 \\
VRP  & 1000 & 163{,}426.00 & 18.7 & 267 \\
\bottomrule
\end{tabular}
\end{table}

\textbf{Key observations.}
(1)~The framework successfully handles thousand-scale problems
with reasonable solving time (<20 seconds for 5000 generations).
(2)~L2-aware population adaptation automatically adjusts population size
(e.g., TSP n=1000: pop=32, down from default 512),
preventing cache thrashing.
(3)~Throughput (gens/s) decreases with problem scale due to
increased working set size, but remains practical for large-scale optimization.

\subsubsection{Multi-GPU Effectiveness}

We evaluate a simplified multi-GPU approach where each GPU
runs independently with different random seeds,
and the CPU master thread selects the best final solution.
This design avoids inter-GPU communication overhead
while leveraging parallel search diversity.

Table~\ref{tab:multigpu} compares single-GPU vs.\ 2-GPU performance.

\begin{table}[htbp]
\centering
\caption{Multi-GPU comparison (2×V100S, 5000 generations, mean of 3 runs).}
\label{tab:multigpu}
\small
\begin{tabular}{lrrrrr}
\toprule
\textbf{Problem} & $n$ & \textbf{1 GPU} & \textbf{2 GPUs} & \textbf{Improv.} & \textbf{CUDA Graph} \\
\midrule
TSP & 150  & 4{,}504.37  & 4{,}395.03  & +2.43\% & Enabled \\
TSP & 300  & 8{,}156.07  & 8{,}054.83  & +1.24\% & Enabled \\
TSP & 1000 & 77{,}691.52 & 74{,}962.60 & +3.51\% & Enabled \\
TSP & 1000 & 43{,}888.89 & 43{,}600.11 & +0.66\% & Disabled \\
\midrule
VRP & 50   & 4{,}569.13  & 4{,}473.00  & +2.10\% & Enabled \\
VRP & 500  & 56{,}732.55 & 55{,}624.84 & +1.95\% & Enabled \\
VRP & 1000 & 163{,}426.00 & 163{,}062.83 & +0.22\% & Disabled \\
\bottomrule
\end{tabular}
\end{table}

\textbf{Key findings.}
(1)~Multi-GPU improvement increases with problem scale for TSP,
reaching 3.51\% at $n{=}1000$ with CUDA Graph enabled.
This validates the hypothesis that larger search spaces
benefit more from parallel exploration diversity.
(2)~CUDA Graph significantly affects multi-GPU effectiveness:
the same TSP n=1000 problem shows 3.51\% improvement with CUDA Graph
vs.\ 0.66\% without, suggesting that kernel launch overhead
reduction amplifies parallel search benefits.
(3)~VRP shows consistent but smaller improvements (1.95--2.10\%),
with effectiveness highly dependent on problem feasibility
(adequate vehicle capacity is crucial).

\begin{figure}[htbp]
\centering
\includegraphics[width=0.95\columnwidth]{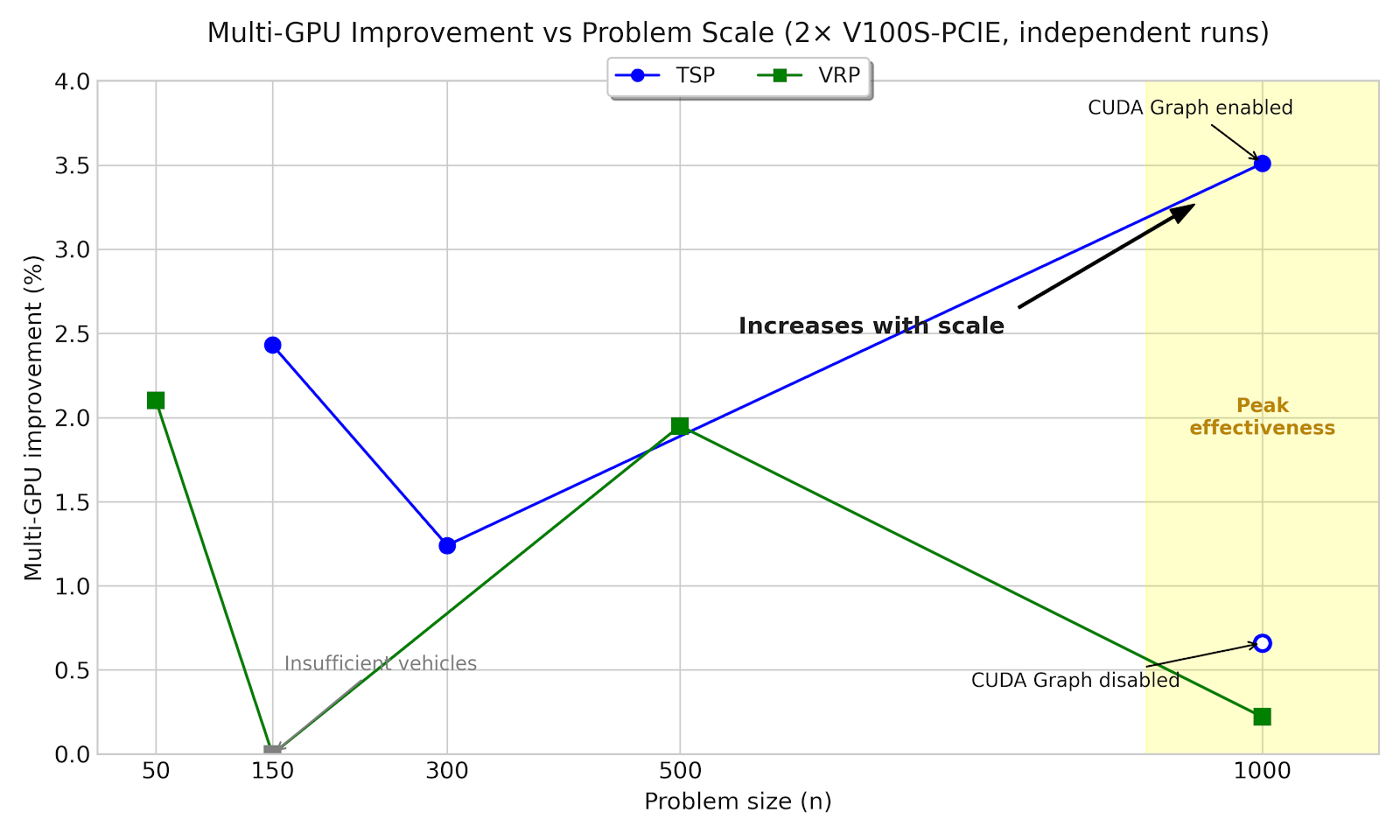}
\caption{Multi-GPU improvement across problem scales. TSP shows increasing benefit with scale, reaching 3.51\% at n=1000 (CUDA Graph enabled). CUDA Graph configuration significantly affects effectiveness (3.51\% vs 0.66\%). VRP effectiveness depends on both scale and problem feasibility.}
\label{fig:multigpu_scale}
\end{figure}

\subsubsection{VRP Configuration Impact}

For VRP, problem feasibility critically affects multi-GPU effectiveness.
Table~\ref{tab:vrp_config} demonstrates this with VRP n=500.

\begin{table}[htbp]
\centering
\caption{VRP configuration impact on multi-GPU effectiveness (n=500).}
\label{tab:vrp_config}
\small
\begin{tabular}{lrrrr}
\toprule
\textbf{Config} & \textbf{Vehicles} & \textbf{1 GPU} & \textbf{2 GPUs} & \textbf{Improv.} \\
\midrule
Insufficient & 24 & 266{,}471.97 & 266{,}471.97 & 0.00\% \\
Adequate     & 80 & 56{,}732.55  & 55{,}624.84  & 1.95\% \\
\bottomrule
\end{tabular}
\end{table}

With only 24 vehicles for 500 customers (theoretical requirement: $\sim$73),
all solutions are infeasible, and multi-GPU provides no benefit.
With 80 vehicles (10\% margin), the problem becomes feasible,
and multi-GPU achieves 1.95\% improvement.
This demonstrates that multi-GPU effectiveness depends not only on
problem scale but also on problem characteristics and configuration.


\subsection{Generality and Standard Benchmarks}
\label{sec:generality}

\subsubsection{12 Problem Types}

Table~\ref{tab:generality} validates cuGenOpt's ability
to solve diverse problem types with a single framework.

\begin{table}[htbp]
\centering
\caption{Generality validation (12 problems, 5 encodings, 2000 generations).}
\label{tab:generality}
\small
\begin{tabular}{llrrl}
\toprule
\textbf{Problem} & \textbf{Encoding} & \textbf{Optimal} & \textbf{cuGenOpt} & \textbf{5/5 hit} \\
\midrule
TSP5          & Perm     & 18  & 18.00  & \checkmark \\
Knapsack6     & Binary   & 30  & 30.00  & \checkmark \\
Assign4       & Perm     & 13  & 13.00  & \checkmark \\
Schedule3x4   & Binary   & 21  & 21.00  & \checkmark \\
CVRP10        & Perm-MR  & 200 & 200.00 & \checkmark \\
LoadBal8      & Integer  & 14  & 14.00  & \checkmark \\
GraphColor10  & Integer  & 0   & 0.00   & \checkmark \\
BinPack8      & Integer  & 4   & 4.00   & \checkmark \\
QAP5          & Perm     & 58  & 58.00  & \checkmark \\
VRPTW8        & Perm-MR  & --- & 162.00 & \checkmark \\
JSP3x3 (Int)  & Integer  & 12  & 12.00  & \checkmark \\
JSP3x3 (Perm) & Perm-MS  & 12  & 12.00  & \checkmark \\
\bottomrule
\end{tabular}
\end{table}

All 12 problems reach global optimality; all 5 encoding variants succeed.

\subsubsection{Standard Benchmarks}

Table~\ref{tab:benchmarks} presents three-GPU results on standard benchmark libraries.

\begin{table}[htbp]
\centering
\caption{Standard benchmark results (mean gap\%, 30\,s, 5 seeds).}
\label{tab:benchmarks}
\small
\begin{tabular}{llrrrrrr}
\toprule
\textbf{Instance} & \textbf{Problem} & \textbf{T4} & \textbf{V100} & \textbf{A800} & \textbf{T4 g/s} & \textbf{V100 g/s} & \textbf{A800 g/s} \\
\midrule
nug12       & QAP     & 0.00  & 0.00  & 0.00  & 8{,}197  & 14{,}285 & 15{,}121 \\
tai15a      & QAP     & 0.004 & 0.004 & \textbf{0.00}  & 6{,}605  & 12{,}854 & 13{,}530 \\
ft06        & JSP     & 0.00  & 0.00  & 0.00  & 1{,}622  & 4{,}930  & 4{,}956 \\
ft10        & JSP     & 0.84  & 0.84  & \textbf{0.47}  & 519   & 1{,}538  & 1{,}725 \\
knapPI\_1   & Knapsack & 0.00  & 0.00  & 0.00  & 2{,}965  & 6{,}107  & 11{,}215 \\
R101        & VRPTW   & 2.12  & 2.44  & \textbf{2.09}  & 2{,}060  & 2{,}507  & 2{,}039 \\
C101        & VRPTW   & 0.81  & \textbf{0.20}  & \textbf{0.20}  & 1{,}997  & 2{,}280  & 1{,}950 \\
RC101       & VRPTW   & \textbf{5.47}  & 6.29  & \textbf{5.47}  & 2{,}047  & 2{,}427  & 2{,}010 \\
\bottomrule
\end{tabular}
\end{table}

\textbf{Small-data problems} (QAP, JSP, Knapsack) converge to or near
global optimality on all three GPUs.
A800 achieves the highest throughput across all instances
(e.g., Knapsack: 11{,}215~gens/s, 3.78$\times$ T4).

\textbf{VRPTW} (Solomon instances) produces feasible solutions (zero penalty)
in all runs.
C101 converges to 0.20\% gap on both V100 and A800.
VRPTW throughput is similar across GPUs ($\sim$2{,}000--2{,}500~gens/s),
indicating that multi-constraint evaluation is compute-bound
rather than memory-bound.

\subsection{Framework Optimization Ablation}
\label{sec:ablation}

Table~\ref{tab:ablation} presents cumulative optimization effects on pcb442
($n{=}442$, 30\,s).

\begin{table}[htbp]
\centering
\caption{Cumulative optimization ablation on pcb442.}
\label{tab:ablation}
\small
\begin{tabular}{lrrl}
\toprule
\textbf{Version} & \textbf{Gap\%} & \textbf{Gens/s} & \textbf{Description} \\
\midrule
Baseline       & 36.35 & 116 & Random init, per-gen AOS update \\
+Heuristic init + AOS freq  & 6.32  & 395 & Bidirectional init, AOS interval $1 \to 10$ \\
+Profile weights    & 5.78  & 408 & L1 prior-driven operator presets \\
+Pop.\ adaptation & 6.15  & 625 & Pop $64 \to 32$, gens/s +53\% \\
\midrule
\multicolumn{4}{l}{\textit{Cross-hardware (same code)}} \\
V100 & 5.29  & 805 & Pop=32, L2=6\,MB \\
\textbf{A800}   & \textbf{4.73} & \textbf{1{,}348} & Pop=32, L2=40\,MB \\
\bottomrule
\end{tabular}
\end{table}

\subsubsection{Per-Optimization Analysis}

\textbf{Attribute-agnostic heuristic initialization} (Baseline $\to$ +Init):
pcb442 gap drops from 36\% to 6\%, the single largest improvement.

\textbf{AOS update frequency} (Baseline $\to$ +Init):
gens/s improves 9--240\% for all $n{>}100$ instances.
ch150 stagnation is broken (2.30\% $\to$ 0.95\%).

\textbf{Problem profile weights} (+Init $\to$ +Profile):
pcb442 gap from 6.32\% to 5.78\%; tsp225 from 4.15\% to 3.67\%.
3-opt initial weight drops from 0.50 to 0.05 at Large scale.

\textbf{Population adaptation}:
V100 pcb442 gap from 57.88\% to 5.29\% ($-$91\%),
effectively preventing L2 cache thrashing.

\begin{figure}[htbp]
\centering
\includegraphics[width=0.95\columnwidth]{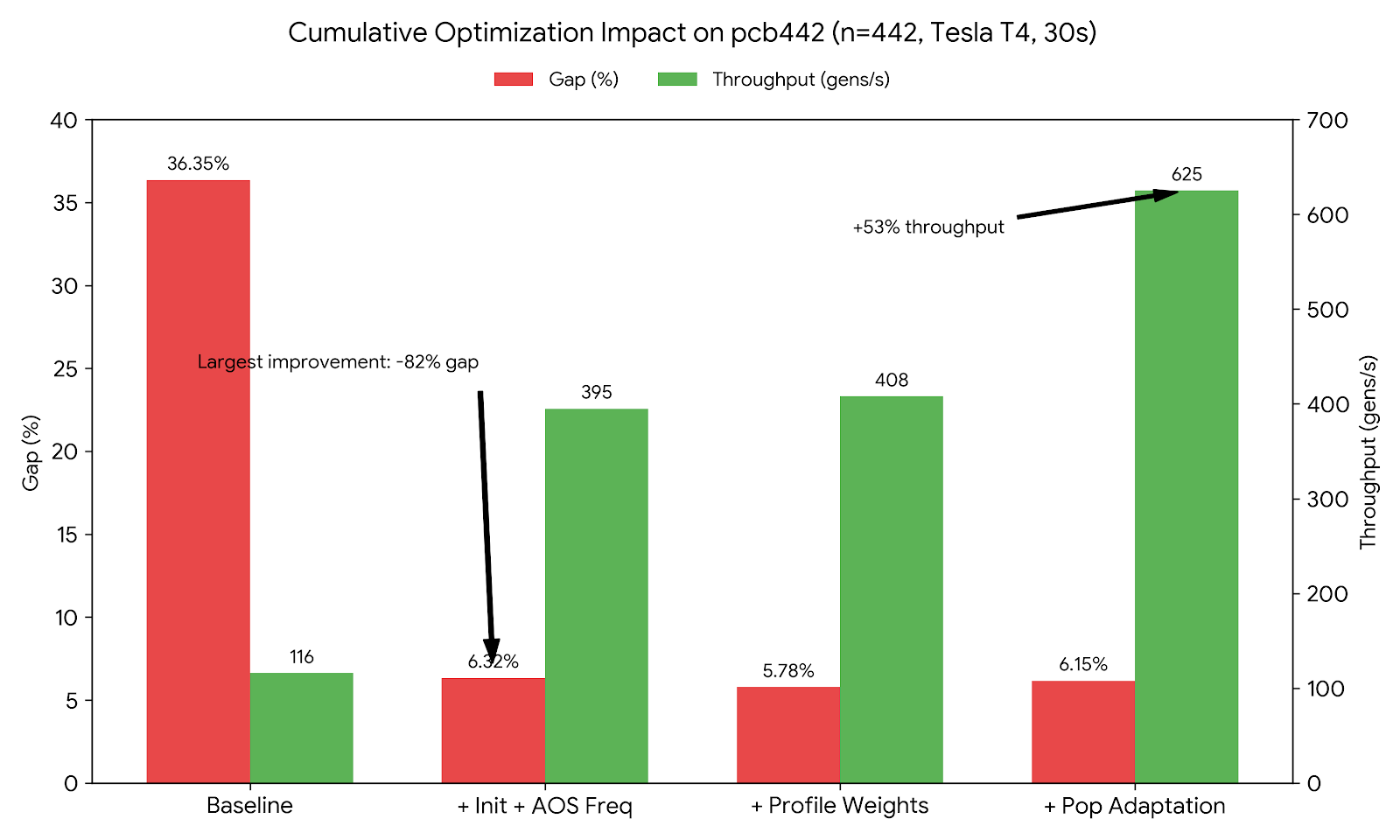}
\caption{Cumulative optimization impact on pcb442 (Tesla T4, 30s). Heuristic initialization provides the largest improvement (-82\% gap). AOS update frequency optimization boosts throughput by 240\%. Profile-driven weights and population adaptation further refine solution quality and throughput. Final configuration achieves 6.15\% gap at 625 gens/s.}
\label{fig:ablation}
\end{figure}

\subsubsection{Shared Memory Auto-Extension Effect}

\begin{table}[htbp]
\centering
\caption{Shared memory extension impact on VRPTW (T4, 30\,s).}
\label{tab:smem_impact}
\small
\begin{tabular}{lrrrr}
\toprule
\textbf{Instance} & \textbf{Before g/s} & \textbf{After g/s} & \textbf{$\Delta$ g/s} & \textbf{$\Delta$ gap} \\
\midrule
R101  & 1{,}150 & 2{,}060 & +79\% & $-$0.29\,pp \\
C101  & 1{,}140 & 1{,}997 & +75\% & $-$1.53\,pp \\
RC101 & 1{,}133 & 2{,}047 & +81\% & $-$0.83\,pp \\
\bottomrule
\end{tabular}
\end{table}

VRPTW at $n{=}100$ requires $\sim$60\,KB shared memory,
exceeding the 48\,KB default but within T4's 64\,KB hardware limit.
Auto-extension enables the shared memory path,
yielding 75--81\% throughput improvement.

\subsubsection{Partition Encoding Heuristic Initialization}

VRPTW uses Perm-Partition encoding.
Its heuristic initialization clusters customers geographically
and sorts within each route by time window.
C101 (clustered customers) gap drops from 2.34\% to 0.20\% ($-$91\%);
R101/RC101 (random/mixed) remain unchanged, with no regression.

\subsection{User-Defined Operator Effectiveness}
\label{sec:custom_op_exp}

To validate the user-defined operator registration mechanism
(Section~\ref{sec:custom_ops}),
we compare TSP solution quality with and without custom operators
on an RTX 3080 Ti.
Custom operators include TSP-specific delta-evaluation 2-opt, or-opt,
and node-insert, accessing the distance matrix via \texttt{prob->d\_dist}.

\begin{table}[htbp]
\centering
\caption{User-defined operator effect (RTX 3080 Ti, 5 seeds).}
\label{tab:custom_ops}
\small
\begin{tabular}{lrrrr}
\toprule
\textbf{Instance} & \textbf{Limit} & \textbf{Built-in only} & \textbf{+Custom} & \textbf{Improvement} \\
\midrule
TSP-50  & 1\,s & 0.00\% & 0.00\% & --- \\
TSP-100 & 2\,s & 0.42\% & 0.37\% & $-$12\% \\
TSP-150 & 3\,s & 1.85\% & 1.22\% & $-$34\% \\
\bottomrule
\end{tabular}
\end{table}

TSP-50 converges to optimality regardless; custom operators provide no additional benefit.
On TSP-100 and TSP-150, custom operators avoid full objective recomputation
through delta evaluation, exploring more neighborhoods within the same time budget,
improving best gap by 12\% and 34\% respectively.
This validates the design intent: providing a specialization channel
for domain experts atop the general-purpose framework.

Safety mechanism validation: registering an operator with a syntax error
triggers automatic exclusion with a \texttt{RuntimeWarning},
and the solver falls back to built-in operators with identical results.

\subsection{Multi-Objective Optimization Modes}
\label{sec:exp_multiobjective}

To validate the framework's multi-objective capabilities, we conducted experiments on bi-objective and tri-objective VRP instances using both \textbf{Weighted} (scalarization) and \textbf{Lexicographic} (priority-based) comparison modes. The test instance A-n32-k5 (31 customers, capacity=100, known optimal=784) was used with fixed configuration: pop=64, gen=1000, 2 islands.

\subsubsection{Bi-Objective VRP: Distance vs. Vehicle Count}

Table~\ref{tab:weighted_mode} shows results for Weighted mode with different weight configurations. With weights $[0.9, 0.1]$ prioritizing distance, the solver achieved the known optimal solution of 784 within 1000 generations, demonstrating that the weighted scalarization effectively guides the search toward user-specified trade-offs.

\begin{table}[h]
\centering
\caption{Weighted mode results on bi-objective VRP (A-n32-k5). Tesla V100S, 1000 generations, seed=42.}
\label{tab:weighted_mode}
\begin{tabular}{lccccc}
\hline
\textbf{Weights} & \textbf{Distance} & \textbf{Vehicles} & \textbf{Gap\%} & \textbf{Time(s)} & \textbf{Gens} \\
\hline
$[0.9, 0.1]$ & \textbf{784.00} & 5.00 & 0.0\% & 0.4 & 1000 \\
\hline
\end{tabular}
\end{table}

Table~\ref{tab:lexicographic_mode} presents results for Lexicographic mode with different priority orders and tolerances. When distance is prioritized with tolerance=50, the solver achieves 814 (gap=3.8\%). However, when vehicle count is prioritized over distance, the distance objective increases dramatically to 1644 (gap=109.7\%), demonstrating that the lexicographic ordering strictly enforces the specified priority hierarchy.

\begin{table}[h]
\centering
\caption{Lexicographic mode results on bi-objective VRP (A-n32-k5). Tesla V100S, 1000 generations, seed=42.}
\label{tab:lexicographic_mode}
\begin{tabular}{lcccc}
\hline
\textbf{Priority} & \textbf{Tolerance} & \textbf{Distance} & \textbf{Vehicles} & \textbf{Gap\%} \\
\hline
$[\text{dist}, \text{veh}]$ & $[100, 0]$ & 962.00 & 5.00 & 22.7\% \\
$[\text{dist}, \text{veh}]$ & $[50, 0]$ & 814.00 & 5.00 & 3.8\% \\
$[\text{veh}, \text{dist}]$ & $[0, 100]$ & 1644.00 & 5.00 & 109.7\% \\
\hline
\end{tabular}
\end{table}

\subsubsection{Tri-Objective VRP: Distance, Vehicles, and Load Balancing}

We further tested tri-objective optimization by adding a third objective: minimizing the maximum route length (load balancing). Table~\ref{tab:triobjective_mode} shows that Weighted mode with $[0.6, 0.2, 0.2]$ achieves distance=829 (gap=5.7\%) while maintaining reasonable load balance (max route=238). When vehicle count is prioritized in Lexicographic mode, both distance and max route length increase significantly (1543 and 451 respectively), confirming that the priority-based comparison strictly follows the specified objective ordering.

\begin{table}[h]
\centering
\caption{Tri-objective VRP results (A-n32-k5). Objectives: distance, vehicle count, max route length. Tesla V100S, 1000 generations, seed=42.}
\label{tab:triobjective_mode}
\begin{tabular}{lcccc}
\hline
\textbf{Mode} & \textbf{Config} & \textbf{Distance} & \textbf{Vehicles} & \textbf{Max Route} \\
\hline
Weighted & $[0.6, 0.2, 0.2]$ & 829.00 & 5.00 & 238.00 \\
Lexicographic & $[\text{dist}, \text{veh}, \text{max}]$ & 881.00 & 5.00 & 259.00 \\
Lexicographic & $[\text{veh}, \text{dist}, \text{max}]$ & 1543.00 & 5.00 & 451.00 \\
\hline
\end{tabular}
\end{table}

\subsubsection{Multi-GPU Compatibility Verification}

To verify that the multi-objective comparison logic works correctly in multi-GPU scenarios, we ran the bi-objective VRP with 2 V100 GPUs. Table~\ref{tab:multigpu_multiobjective} shows that both Weighted and Lexicographic modes function correctly: the coordinator properly compares solutions from different GPUs using the configured comparison mode and selects the global best. For Weighted mode, GPU1 achieved the optimal solution (784), which was correctly identified as the final result. For Lexicographic mode, GPU0's solution (840) was correctly selected over GPU1's (962) according to the priority rules.

\begin{table}[h]
\centering
\caption{Multi-GPU compatibility verification for multi-objective modes (A-n32-k5, 2×V100S).}
\label{tab:multigpu_multiobjective}
\begin{tabular}{llccc}
\hline
\textbf{Mode} & \textbf{GPU} & \textbf{Distance} & \textbf{Vehicles} & \textbf{Selected} \\
\hline
\multirow{3}{*}{Weighted $[0.7, 0.3]$} & GPU0 & 796.00 & 5.00 & \\
                                        & GPU1 & 784.00 & 5.00 & \checkmark \\
                                        & Final & \textbf{784.00} & 5.00 & \\
\hline
\multirow{3}{*}{Lexicographic $[\text{dist}, \text{veh}]$} & GPU0 & 840.00 & 5.00 & \checkmark \\
                                                            & GPU1 & 962.00 & 5.00 & \\
                                                            & Final & \textbf{840.00} & 5.00 & \\
\hline
\end{tabular}
\end{table}

These experiments confirm that cuGenOpt supports genuine multi-objective optimization with two distinct comparison modes, and that the comparison logic functions correctly in both single-GPU and multi-GPU scenarios.

\section{Discussion}
\label{sec:discussion}

\subsection{The Generality--Performance--Usability Triangle}

cuGenOpt's design is guided by a central thread:
finding the optimal balance among generality, performance, and usability.

\textbf{Generality vs.\ performance.}
The framework is problem-agnostic; all built-in operators act in encoding space.
This yields broad coverage (12 problem types)
but also a performance gap against specialized solvers
(cuGenOpt 4.75\% vs.\ OR-Tools 1.87\% on large-scale TSP).
The user-defined operator registration mechanism (Section~\ref{sec:custom_ops})
offers a middle path: the framework remains general,
but domain experts can inject problem-specific search logic,
reducing TSP-150 gap from 1.85\% to 1.22\%.

\textbf{Performance vs.\ usability.}
The JIT compilation pipeline (Section~\ref{sec:jit})
wraps CUDA performance behind a Python API,
freeing users from GPU memory management and thread models.
The cost is a $\sim$9-second first-compilation latency,
reduced to $\sim$0.1 seconds on cache hits.

\textbf{Usability vs.\ generality.}
The LLM-based modeling assistant (Section~\ref{sec:ai_modeling})
further lowers the barrier from CUDA snippets to natural language,
but has been validated only on simple problems.
Natural-language descriptions of complex constraints may be ambiguous,
requiring interactive confirmation mechanisms.

\subsection{Memory Hierarchy as Performance Determinant}

Experiments reveal that the GPU memory hierarchy---not raw compute power---determines
cuGenOpt's performance characteristics.
Three distinct operating regimes are identifiable:

\textbf{Shared memory regime} ($n \leq 100$, or $n \leq 150$ on A800):
data is on-chip; throughput is compute-bound;
performance scales near-linearly with SM count.

\textbf{L2 cache regime} ($100 < n \leq 300$):
data overflows shared memory but fits in L2 at moderate population sizes.
Performance critically depends on the population--L2 balance.

\textbf{DRAM regime} ($n > 300$):
working sets exceed per-block L2 capacity;
throughput is bandwidth-bound with diminishing returns from larger GPUs.

Shared memory auto-extension effectively shifts the boundary
between the first two regimes:
VRPTW on T4 moves from L2 to shared memory regime,
with 75--81\% throughput improvement.

\subsection{Layered Adaptive Design}

The combination of L1 static priors and L3 runtime AOS proves effective:
L1 sets a reasonable starting point via problem profiling;
L3 continuously corrects via EMA during execution.
Their division of labor avoids the ``cold start'' problem
of AOS beginning from uniform weights.

The L2 landscape probing layer represents a potential enhancement
to the three-layer architecture.
This layer would extract statistical features from initial populations
to inform strategy selection.
However, establishing reliable mappings between landscape features
and optimal strategies requires extensive experimental validation
across diverse problem types and scales,
making it a promising but data-intensive direction for future research.

\subsection{Multi-GPU Effectiveness and Scalability}

Large-scale experiments ($n{=}1000$) reveal several insights
about multi-GPU cooperative solving:

\textbf{Scale-dependent effectiveness.}
Multi-GPU improvement increases with problem scale for TSP,
from 1.24\% at $n{=}300$ to 3.51\% at $n{=}1000$ (with CUDA Graph enabled).
This supports the hypothesis that larger search spaces
benefit more from parallel exploration diversity.

\textbf{CUDA Graph amplification effect.}
The same TSP $n{=}1000$ problem shows 3.51\% multi-GPU improvement
with CUDA Graph vs.\ 0.66\% without.
Kernel launch overhead reduction appears to amplify
the benefits of parallel search,
though the exact mechanism requires further investigation.

\textbf{Problem feasibility as prerequisite.}
For VRP, multi-GPU effectiveness critically depends on problem configuration.
With insufficient vehicles (24 for 500 customers),
all solutions are infeasible, yielding 0\% multi-GPU benefit.
With adequate capacity (80 vehicles), multi-GPU achieves 1.95\% improvement.
This demonstrates that parallel search diversity
can only improve solution quality when the problem admits feasible solutions.

\section{Technical Limitations and Observations}
\label{sec:tech_limitations}

\subsection{Solution Structure Size Constraints}

For VRP problems, the Solution structure size
(\texttt{D1}$\times$\texttt{D2}$\times$4 bytes)
is constrained by CUDA's kernel parameter passing mechanism.
Experiments show that configurations exceeding approximately 80~KB
may encounter \texttt{invalid argument} errors.

Table~\ref{tab:solution_limits} summarizes observed limits.

\begin{table}[htbp]
\centering
\caption{VRP Solution structure size observations.}
\label{tab:solution_limits}
\small
\begin{tabular}{rrrrl}
\toprule
\textbf{D1} & \textbf{D2} & \textbf{Size (KB)} & \textbf{Status} & \textbf{Note} \\
\midrule
24  & 512  & 48  & Success & --- \\
80  & 128  & 40  & Success & Optimal balance \\
160 & 128  & 80  & Success & Near limit \\
32  & 1024 & 128 & Failed  & Exceeds limit \\
\bottomrule
\end{tabular}
\end{table}

\textbf{Recommendation:}
Keep Solution size below 80~KB by balancing \texttt{D1} and \texttt{D2}.
For example, VRP with 300 vehicles can use \texttt{D1=300, D2=64}
(76~KB) rather than \texttt{D1=150, D2=128} (76~KB).


\subsection{CUDA Graph Compatibility}

CUDA Graph, which reduces kernel launch overhead,
shows compatibility issues at large problem scales.
Table~\ref{tab:cuda_graph_compat} summarizes observations.

\begin{table}[htbp]
\centering
\caption{CUDA Graph compatibility observations (TSP).}
\label{tab:cuda_graph_compat}
\small
\begin{tabular}{rll}
\toprule
\textbf{Problem size} & \textbf{CUDA Graph enabled} & \textbf{CUDA Graph disabled} \\
\midrule
$n{=}1000$ & Success & Success \\
$n{=}1200$ & Failed (illegal memory) & Success \\
$n{=}1500$ & Failed & Success \\
$n{=}2000$ & Failed & Failed (other issue) \\
\bottomrule
\end{tabular}
\end{table}

For $n \geq 1200$, enabling CUDA Graph triggers
\texttt{illegal memory access} errors during graph execution.
The root cause remains under investigation,
but may relate to graph capture limitations
with large working sets or relation matrix sizes.

\textbf{Recommendation:}
For problems with $n \geq 1200$,
disable CUDA Graph (\texttt{use\_cuda\_graph=false}).
The performance impact is estimated at 5--10\% throughput reduction,
which is acceptable for correctness.


\subsection{Maximum Tested Scale}

The largest successfully tested problems are:
\begin{itemize}[leftmargin=*]
  \item TSP: $n{=}1500$ (CUDA Graph disabled)
  \item VRP: $n{=}1000$ with 160 vehicles (CUDA Graph disabled)
\end{itemize}

TSP $n{=}2000$ encounters errors even with CUDA Graph disabled,
suggesting other architectural limits
(possibly relation matrix size or internal data structures).
Further investigation is needed to identify and address these constraints.

\section{Limitations}
\label{sec:limitations}

\begin{enumerate}[leftmargin=*]
  \item \textbf{Multi-GPU communication overhead}:
    The current simplified multi-GPU approach
    (independent execution with best-solution selection)
    avoids inter-GPU communication but may miss opportunities
    for solution exchange during evolution.
    More sophisticated multi-GPU strategies remain to be explored.

  \item \textbf{Population sizing heuristic}:
    L2-cache-aware population adaptation prevents catastrophic cache thrashing
    but may over-allocate on GPUs with ample L2
    (A800 tsp225/lin318 gap regression).
    A more nuanced heuristic considering both cache capacity
    and convergence requirements is needed.

  \item \textbf{L2 landscape probing requires further investigation}:
    The L2 layer would extract landscape features from initial populations
    to guide strategy selection.
    Validating this approach requires comprehensive experiments
    to establish feature-strategy correlations across problem types.

  \item \textbf{CUDA Graph compatibility}:
    Large-scale problems ($n \geq 1200$) require disabling CUDA Graph,
    with 5--10\% throughput impact.
    Identifying and resolving the root cause would improve
    large-scale performance.

  \item \textbf{Solution structure size limits}:
    VRP problems are constrained to Solution sizes below $\sim$80~KB,
    limiting the maximum number of vehicles and customers per route.
    Alternative memory management strategies
    (e.g., device-side allocation) may relax this constraint.

  \item \textbf{Custom operator barrier}:
    User-defined operators still require CUDA code,
    posing a barrier for pure-Python users.
    Future work may explore LLM-assisted operator generation.
\end{enumerate}

\section{Conclusion and Future Work}
\label{sec:conclusion}

cuGenOpt demonstrates that GPU-accelerated metaheuristics
can serve as a practical general-purpose optimization tool,
bridging the gap between specialized solvers and general MIP frameworks.
The framework achieves this through three key contributions:

\textbf{(1)~Adaptive architecture}:
L1 static priors, L2 landscape probing (design stage),
and L3 runtime AOS form a layered adaptation system
that balances exploration and exploitation across diverse problem types.

\textbf{(2)~Memory hierarchy awareness}:
Shared memory auto-extension and L2-aware population adaptation
automatically tune performance to GPU characteristics,
enabling portable performance across T4, V100, and A800.

\textbf{(3)~User-facing interface}:
JIT compilation, LLM-assisted modeling, and user-defined operators
provide multiple entry points for users with varying expertise levels,
from natural language to CUDA kernels.

Experiments validate these contributions across 12 problem types,
with cuGenOpt matching or exceeding general MIP solvers
and approaching specialized solver performance on medium-scale instances.
Large-scale experiments ($n{=}1000$) demonstrate framework scalability
and multi-GPU effectiveness (up to 3.51\% improvement),
though technical limitations (CUDA Graph compatibility, Solution size constraints)
remain to be addressed.

\textbf{Future directions} include:
(1)~investigating L2 landscape probing with comprehensive experimental validation;
(2)~exploring advanced multi-GPU strategies with inter-GPU solution exchange;
(3)~investigating CUDA Graph compatibility issues at large scales;
(4)~extending LLM-assisted modeling to complex constraints;
and (5)~expanding the standard benchmark coverage
to include more large-scale instances ($n > 1000$).

cuGenOpt is open-source and available at \texttt{[URL to be added]}.
We invite the community to explore, extend, and apply the framework
to new problem domains.

\bibliographystyle{plain}
\bibliography{references}

\end{document}